\documentclass{article}

\usepackage[final]{corl_2020} 
\pdfoutput=1
\usepackage{cite}
\usepackage{amsmath,amssymb,amsfonts}
\usepackage{graphicx}
\usepackage{hyperref}
\usepackage{textcomp}
\usepackage{xcolor}
\usepackage{xspace, footmisc}
\usepackage{tikz}
\usepackage{pgfplots}
\usetikzlibrary{arrows,backgrounds,decorations,decorations.pathmorphing,positioning,fit,automata,shapes,patterns,plotmarks,calc}
\usepackage{subfig}
\usepackage{stmaryrd}
\usepackage[inline]{enumitem}
\usepackage{booktabs}
\usepackage{todonotes}
\graphicspath{{figs/}}
\usepackage{wrapfig}
\usepackage{macros}
\usepackage{multirow}
\usepackage{wrapfig}
\usepackage[ruled, vlined]{algorithm2e}
\usepackage{multicol, nccmath}

\newcommand{\qstl}{$Q_{stl}$ }

\title{Learning from Demonstrations using \\ Signal Temporal Logic}

\author{
  Aniruddh G.~Puranic, Jyotirmoy V.~Deshmukh and Stefanos Nikolaidis\\
  University of Southern California \\
  \texttt{\{puranic, jdeshmuk, nikolaid\}@usc.edu} \\
}

\begin{document}
\maketitle

\vspace*{-20pt}
\begin{abstract}
Learning-from-demonstrations is an emerging paradigm to obtain effective robot control policies for complex tasks via reinforcement learning without the need to explicitly design reward functions. However, it is susceptible to imperfections in demonstrations and also raises concerns of safety and interpretability in the learned control policies. To address these issues, we use Signal Temporal Logic to evaluate and rank the quality of demonstrations. Temporal logic-based specifications allow us to create non-Markovian rewards, and also define interesting causal dependencies between tasks such as sequential task specifications. We validate our approach through experiments on discrete-world and OpenAI Gym environments, and show that our approach outperforms the state-of-the-art Maximum Causal Entropy Inverse Reinforcement Learning.
\end{abstract}



\section{Introduction}
\vspace*{-10pt}
One of the emerging methods to design control policies for robots is the paradigm of {\em learning-from-demonstrations} (LfD) \citep{lfd_schaal1, lfd_schaal2}. This paradigm has led to vibrant research on a number of different approaches such as apprenticeship learning (AL) \citep{abbeel_ng}, inverse reinforcement learning (IRL) \citep{Ng_russell, ziebart_maximumentropy}, and behavior cloning via supervised learning \citep{bco_stone}. IRL seeks to recover the reward function from a set of human demonstrations that could be generalized to similar reinforcement learning (RL) tasks. Behavior cloning on the other hand relies on supervised learning to model/mimic the actions of a teacher. Designing reward functions for RL tasks requires expert knowledge in this domain and is not trivial to recover rewards \citep{ravichandar_recent_2020}. In addition, it is difficult even for experts to design reward functions for RL tasks that involve multiple and/or sequential objectives. 

At its core, LfD provides a mechanism to indirectly provide specifications on expected behavior of a robot, and learning a control policy from these specifications. LfD can also address the issue in designing rewards for multiple objectives. However, there are methodological limitations to the prevalent LfD paradigm: (i) a demonstration is an inherently incomplete and implicit specification of the robot behavior in a specific fixed initial configuration or in the presence of a single disturbance profile. The control policy that is inferred from a demonstration may thus perform unsafe or undesirable actions when the initial configuration or disturbance profile is different \citep{hussein_imitation_2017}. Thus, learning from demonstrations lacks robustness, (ii) not all demonstrations are equal: some demonstrations are a better indicator of the desired behavior than others, and the quality of a demonstration often depends on the expertise of the user providing the demonstration \citep{ravichandar_recent_2020}. There is also lack of metrics to evaluate the quality of demonstrations on tasks \citep{osa_algorithmic_2018, hussein_imitation_2017}, (iii) demonstrations have no way of explicitly specifying safety conditions for the robot, and safely providing a demonstration is itself a skill \citep{hussein_imitation_2017, ravichandar_recent_2020}, (iv) there may be many optimal demonstrations, each trying to optimize a particular objective (also known as user preference).

In order to overcome some of these shortcomings, we propose a technique where the user provides partial specifications in a mathematically precise and unambiguous formal language. In this work,we use the formalism of {\em Signal Temporal Logic} (STL) as the specification language of choice, but our framework is flexible to allow other kinds of formalisms. In recent years, STL has been extensively used in cyber-physical system applications~\citep{donze_automotive_2015,kapinski_st-lib_2016,bartocci_specification-based_2018}. Essentially a formula in STL is evaluated over a temporal behavior of the system (e.g. a multi-dimensional signal consisting of the robot's position, joint angles, angular velocities, linear velocity etc.). STL allows Boolean satisfaction semantics: a behavior satisfies a given formula or violates it. A more useful feature of STL in the context of our work is its quantitative semantics that define how {\em robustly} a signal satisfies a formula or define a {\em signed distance} of the signal to the set of signals satisfying the given formula~\citep{deshmukh_robust_2017,fainekos_robustness_2009}. 

Certain assumptions about the $task$ and $environment$ are required to learn accurate cost or reward functions and thus cannot be generalized to other applications without modification \citep{ravichandar_recent_2020}, and STL is one of the ways of defining properties of tasks and environments. We use STL specifications for two distinct purposes: (1) to evaluate and automatically rank demonstrations based on their fitness w.r.t. the specifications, and (2) to infer rewards to be used in a {\em model-free RL} procedure used to train the control policy. The quality of demonstrations, also known as fitness, is the degree of satisfaction of the demonstration on the defined STL specifications. We present a novel way of estimating the quality of a demonstration over a set of specifications by representing the specifications in a directed acyclic graph to encode the relative priorities among them.

An important problem to address when designing and training RL agents is the design of {\em state-based reward functions} \citep{sutton_reinforcement_2018} as a means to incorporate knowledge of the goal and the environment model in training an RL agent. As reward functions are mostly handcrafted and tuned, poorly designed reward functions can lead to the RL algorithm learning a policy that produces undesirable or unsafe behaviors or simply to a task that remains incomplete \citep{leike_ai_2017}. {\em The key insight of this work is that the use of even partial STL specifications can help in a mechanism to automatically evaluate and rank demonstrations, leading to learning robust control policies and inferring rewards to be used in a model-free RL setting.} The ultimate objective of this work is to provide a framework for a flexible structured reward function formulation. The main contributions of our work are:
\begin{enumerate}[topsep=0pt,itemsep=-1ex,partopsep=1ex,parsep=1ex,leftmargin=1.5em]
\item We propose a framework for LfD using STL specifications to infer rewards without the necessity for optimal or perfect demonstrations. In other words, our method can infer non-Markovian rewards even from imperfect or sub-optimal demonstrations and are used by the robot to find a policy using off-the-shelf model-free RL algorithms with slight modifications.
\item We show that our method can also learn from only a small number of demonstrations which is practical for non-expert users and also for large environments that result in sparse rewards, while not introducing additional hyperparameters for the reward inference procedure.
\item We also tackle the problem of achieving multiple sequential goals/objectives by combining STL specifications with Q-Learning. Using a discrete-world setting, we show that effective control policies can be learned such that they satisfy the defined safety requirements while also trying to imitate the user preferences.
\end{enumerate}


\vspace*{-10pt}
\section{Preliminaries}
\vspace*{-5pt}
\begin{definition}[Environment]
	It is a tuple  $E = (S, A)$ consisting of the set of all possible states $S$ defined over $\mathbb{R}^{n}$ and actions $A$, where $n$ is the dimension of the real space. A goal or objective in $E$ is an element of $S$. 
	\label{def:env}
\end{definition}
\vspace*{-5pt}
\begin{definition}[Demonstration]
	A demonstration (or a policy or trace) is a finite sequence of state-action pairs. Formally, a demonstration $d$ of length $L \in \mathbb{N}$ is given as $d =\{(s_1, a_1), (s_2, a_2), ..., (s_L, a_L)\}$, where $s_i \in S$ and $a_i \in A$. That is, $d$ is an element of $(S \times A)^L$. \label{def:demo}
\end{definition}



\textit{Signal Temporal Logic (STL)} is a real-time logic, generally interpreted
over a dense-time domain for signals that take values in a continuous metric
space (such as $\Reals^m$). For a policy or demonstration, the basic primitive in STL is a {\em signal predicate} $\mu$ that is a formula of the form $f(\vx(t)) > 0$, where $\vx(t)$
is the tuple $(state, action)$ of the demonstration $\vx$ at time $t$, and $f$ is a function from the signal domain $\domain = (S \times A)$ to $\Reals$. STL formulas are then defined recursively using Boolean combinations of sub-formulas, or by applying an
interval-restricted temporal operator to a sub-formula.  The syntax of STL is
formally defined as follows: $\varphi ::=  \mu \mid \neg \varphi \mid \varphi \wedge \varphi          \mid \alw_{I} \varphi  \mid \ev_{I} \varphi \mid \varphi \until_{I} \varphi$.
Here, $I = [a,b]$ denotes an arbitrary time-interval, where $a,b\in\Reals^{\ge
0}$. The semantics of STL are defined over a discrete-time signal $\sig$ defined
over some time-domain $\timedomain$. The Boolean satisfaction of a signal predicate
is simply \textit{True} ($\top$) if the predicate is satisfied and \textit{False} ($\bot$) if it is not, the semantics for the propositional logic operators $\neg, \land$ (and thus $\lor,
\rightarrow$) follow the obvious semantics. The temporal operators model the following behavior:
\begin{itemize}[leftmargin=1em,nosep]
    \item At any time $t$, $\always_I(\varphi)$ says that $\varphi$ must hold for all samples in $t+I$.
    \item At any time $t$, $\eventually_I(\varphi)$ says that $\varphi$ must hold \textit{at
    least once} for samples in $t+I$.
    \item At any time $t$, $\varphi \until_I \psi$ says that $\psi$ must hold at time $t’$ in $t+I$, and in $[t,t’)$, $\varphi$ must hold.
\end{itemize}


A signal satisfies an STL formula $\varphi$ if it is satisfied at time $t=0$. The quantitative semantics of STL are defined in the appendix. Intuitively, they represent the numerical distance of ``how far" a signal is away from the signal predicate. For a given requirement $\varphi$, a demonstration or policy $d$ that satisfies it is represented as $d \models \varphi$ and one that doesn't is represented as $d \not\models \varphi$.

\begin{wrapfigure}[21]{r}{.35\textwidth}
      \centering
      \begin{minipage}{\linewidth}
      \centering
        \includegraphics[scale=0.3]{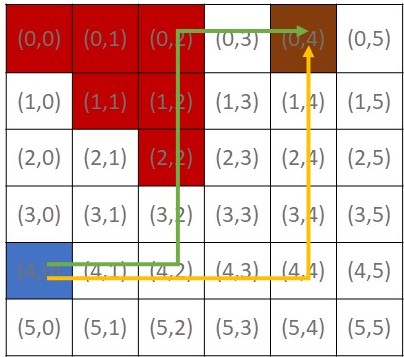}
        \caption{Demonstrations in a grid-world. \label{fig:stl_demo_eg}}
      \end{minipage}
      \hfill
      \begin{minipage}{\linewidth}
        \centering
        \includegraphics[scale=0.25]{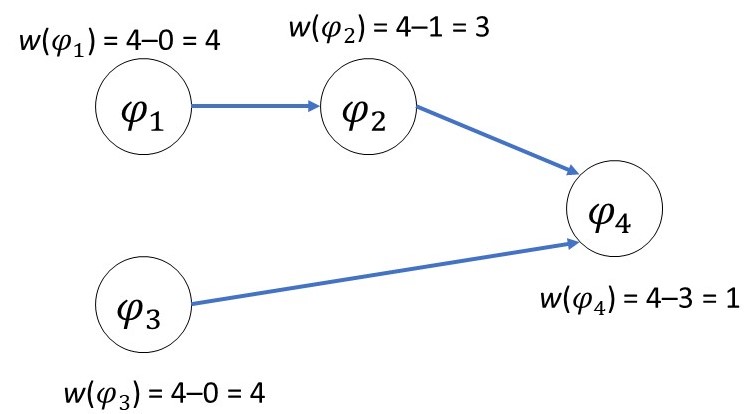}
        \caption{Weights on nodes in a DAG. \label{fig:dag_eg}}
      \end{minipage}
\end{wrapfigure}

\begin{example}
    Consider a $6 \times 6$ grid environment and the policies shown in $green$ and $yellow$ (\autoref{fig:stl_demo_eg}). Each cell in the grid is represented a tuple (x, y) indicating its coordinates with the origin at top-left $(0,0)$. The possible actions in each cell are $\{U, D, L, R\}$. The red cells are regions to be avoided and a policy is required to start at blue cell and end at brown. Consider the specifications: $\varphi_1 := \ev_{[0, 9]}(x(t) = (0,4))$ and $\varphi_2 := \alw_{[0,9]}(dist_\mathrm{red}(x(t)) \ge 1)$ where $dist_\mathrm{red}$ is the taxi-cab distance between a cell and its nearest red cell. For the $green$ policy, $\pi = \{((4,0),R);((4,1),R);((4,2),U);((3,2),U);((2,2),U);((1,2),U);((0,2),R);((0,3),R);\\((0,4),U)\}$. Here we consider the signal $\vx$ to represent only the states of $\pi$. We see that $\varphi_1$ is satisfied since the brown cell $(0,4)$ occurs in the policy within 9 time-steps. We compute the $dist_\mathrm{red}$ and we see that the policy intersects with red cells and hence $\varphi_2$ is not satisfied since there exists a time-step at which the cells coincide. In a similar way, we can see that the $yellow$ policy satisfies both requirements since the goal state occurs in its policy and its $dist_\mathrm{red}$ is always greater than 0.
    
	

\end{example}

	  

\vspace*{-5pt}
There are two classes of temporal logic requirements: (i) hard requirements $\Phi_H$ and (ii) soft requirements $\Phi_S$. Hard requirements are the certain properties of a system that are required to be invariant, i.e., the system must obey the rules or operate within its constraints at all times. Examples of this are: a robot should always operate/remain within its operational workspace, the joint velocities of a robot must always be within a specific range $[v_a, v_b]$, etc. These properties can be interpreted as safety requirements for the system and they typically have the form: $\alw(\varphi)$. Such requirements always need to be satisfied by a system before being able to satisfy the soft requirements. Soft requirements typically correspond to the optimality of a system such as performance, efficiency, etc. These specifications may also be competing with each other and might require some trade-offs.

\vspace*{-15pt}
\section{Methodology}
\vspace*{-15pt}
\mypara{Problem Formulation} In this work, we aim to infer rewards from user demonstrations and STL specifications. Given a transition system  $M \backslash \{R, T\}$ with unknown rewards and transition probabilities, a finite set of high-level specifications in STL $\Phi = \Phi_H \cup \Phi_S$ and a finite dataset of human demonstrations $D = \{d_1, d_2, ..., d_m\}$ in an environment $E$, where each demonstration is defined as in \autoref{def:demo}, the goal is to infer a reward function $R$ for $M$ such that the resulting robot policy $\pi$ obtained by a model-free RL algorithm, satisfies all the requirements of $\Phi$ \footnote{The ideal procedure would involve verification, but we just empirically verify. \label{fnlabel}}. The hard requirements are given by $\Phi_H = \{\varphi_1, \varphi_2, ..., \varphi_p\}$ and the soft requirements are given by  $\Phi_S = \{\varphi_{p+1}, \varphi_{p+2}, ..., \varphi_{q}\}$.
\vspace*{-5pt}

\begin{wrapfigure}[12]{l}{.6\textwidth}
    \begin{minipage}[tb]{\linewidth}
    \centering
\includegraphics[width=\linewidth]{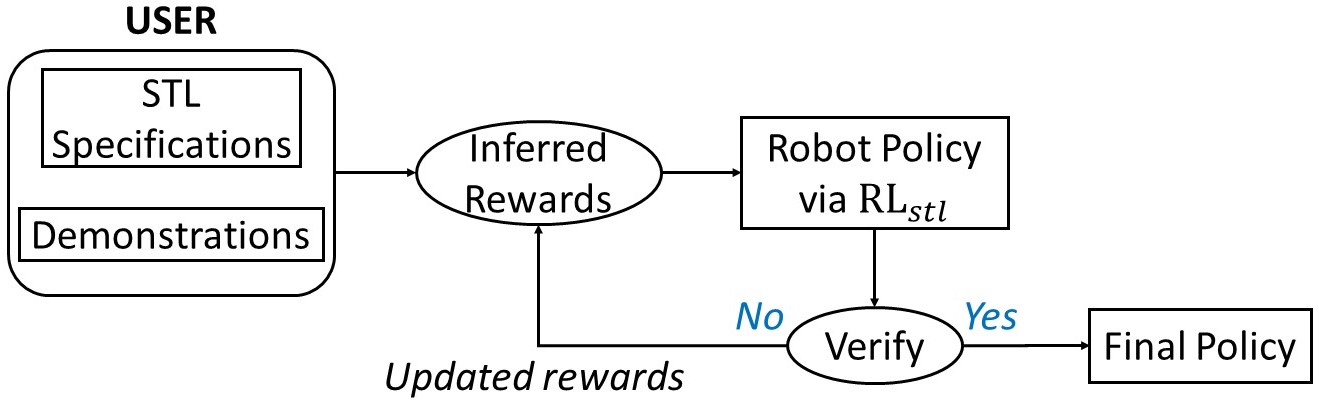}
    \end{minipage}
    \caption{Framework for integrating LfD and STL to infer reward functions and robot policy.}
    \label{fig:framework}
\end{wrapfigure}

\mypara{Framework}
In this section we design a framework (\autoref{fig:framework}) for learning reward functions from demonstrations and STL specifications. A user defines a set of specifications or system requirements which are arranged in a directed acyclic graph structure, explained in \autoref{sec:dag}. The user also provides a demonstration-set $D$, which is then utilized by algorithms described in \autoref{sec:dag}, to infer a reward function $R$ for the robot. The problem is then solved through
 a feedback loop (\autoref{fig:framework}) on the inferred reward $R$ using our proposed model-free RL algorithm (\autoref{sec:rl_stl}), to obtain a robot policy that satisfies the user requirements \footref{fnlabel}. Using the STL representations, we can express complex tasks involving multiple goals, which cannot be easily encoded or represented in traditional IRL. The assumptions in this work are: (1) The world and agent consist of discrete states and actions \footnote{For continuous state systems, we perform an abstraction that groups several continuous states into abstract discrete states to avoid the curse of dimensionality.}, (2) we assume that there exists a feasible path to reach the goals from the initial state, (3) for testing on an unseen map, we only require that the map is of the same size as the one on which the robot was trained. We also consider only the states of a policy as our signal and discard the actions associated with those states when evaluating a specification.

\vspace*{-10pt}
\subsection{Reward Inference}
\label{sec:dag}
\vspace*{-12pt}
\mypara{DAG Representation}
A \textit{Directed Acyclic Graph (DAG)} is an ordered pair $G = (V, E)$ where $V$ is a set of elements called vertices or nodes and $E$ is a set of ordered pairs of vertices called edges or arcs, which are directed from one vertex to another. An edge $e = (u, v)$ is directed from vertex $u$ to vertex $v$. A path $p(u, v)$ in $G$ is a set of vertices starting from $u$ and ending at $v$ by following the directed edges from $u$. The ancestors of a vertex $v$ is the set of all vertices in $G$ that have a path to $v$. Formally, $ancestor(v) = \{u \mid p(u, v), u \in V\}$. The requirements in $\Phi_H$ and $\Phi_S$ are each represented as a node in a DAG $G$ since our intention is to explicitly capture dependencies between requirements: we need requirements in $\Phi_H$ to be satisfied before the requirements in $\Phi_S$ are satisfied. Thus, edges in the DAG capture dependencies and user preferences among requirements. The weight on each node in $G$ is computed using \autoref{eqn:dag_weights} and an example is shown in \autoref{fig:dag_eg}.
\useshortskip\begin{equation}
\label{eqn:dag_weights}
w(\varphi) = |\Phi| - |ancestor(\varphi)|
\end{equation}
where $\Phi = \Phi_H \cup \Phi_S$ is the set of all specifications. This equation represents the relative importance of each specification based on the number of dependencies that need to be satisfied. These computed weights are passed through a $\mathtt{softmax}$ function to give higher importance to ``harder" specifications. For an STL specification $\varphi_i \in \Phi$ and a demonstration $d_j \in D$ defined as in \autoref{def:demo}, the value $rob_i^j = \rho(\varphi_i, d_j, t)$ represents how well the demonstration satisfies the given specification, i.e., the robustness value is used to assess quality of the demonstration w.r.t the specification. There are two reward inference rules based on the quality of a demonstration. At a given time $t$ and for every demonstration $d_j \in D$, the final reward is computed as in \autoref{eqn:total_demo_rob}, where $q$ is the total number of specifications in $\Phi$ of which the first $p$ are $\Phi_H$ and the remaining $q-p$ are $\Phi_S$. The reward $r_{d_j} \in R$ where $R:D \rightarrow \mathbb{R}$, i.e., it maps a demonstration to a real number. 
    
\noindent\begin{minipage}{.5\linewidth}
\begin{equation}
    r_{d_j} = \sum_{i=1}^{q} w(\varphi_i) \cdot rob_i^j
    \label{eqn:total_demo_rob}
    \end{equation}
\end{minipage}%
\begin{minipage}{.5\linewidth}
    \begin{equation}
    \label{eqn:bad_reward}
    r(s) =
        \begin{cases}
          r_{d_j}, & \text{if}\ s \in s_{bad} \\
          0, & \text{otherwise}
        \end{cases}
    \end{equation}
\end{minipage}

In addition, the robustness values can be bounded to specific ranges depending on the STL formula, such as using $\mathtt{tanh}$ or piece-wise linear functions. This makes it appropriate to linearly combine robustness values of specifications since they are on similar scales. For a demonstration, the rewards in each state must be assigned a numerical value based on $r_{d_j}$ described in the following sections. The rewards for $d_j$ are $\{r(s_1), r(s_2), ..., r(s_L)\}$ where $r(s)$ is the reward corresponding to each state $s \in d_j$. 

\mypara{Specification-ranked demonstrations}
\begin{definition}[Good demonstration] We call a demonstration {\em good} if the sequence of state-action pairs in the demonstration satisfies all STL requirements. Every state or state-action pair of the demonstration does not violate any specification. 
\end{definition}
Based on this reasoning, the reward is assigned to every state in the demonstration, while other states are assigned a reward of zero. Let a demonstration $d_j$ of length $L$ have a reward value $r_{d_j}$ computed using \autoref{eqn:total_demo_rob}. The reward assignment capturing the \textit{non-Markovian} or cumulative nature is given as: $\mathbf{r(s_l) = \frac{l}{L} \cdot r_{d_j}}$, where $l = 1, 2, ... L; s_l \in d_j$. This essentially captures the \textit{non-Markovian} nature of the demonstration since the entire trajectory is evaluated, and based on the above equation, the reward at each step guides the robot towards the goal along the demonstrated path. The good demonstrations will have strictly non-negative robustness value and hence positive rewards.

\begin{definition}[Bad Demonstration]
A ``bad" demonstration is one which does not satisfy any of the hard STL requirements $\Phi_H$. The demonstration may be imperfect, incomplete or both. At least one state-action pair in the demonstration fails to satisfy any of hard STL requirements. Mathematically, given a hard requirement $\varphi$ of the form $\alw(\psi)$, a demonstration is bad if $\exists j: s.t. (s_j, a_j)  \not\models \psi$.
\end{definition}

Logically, instead of assigning rewards to each state of the demonstration, the reward is only assigned to the states or state-action pairs violating the specifications, while other states are assigned a reward of zero. A bad demonstration will have non-positive robustness value and hence negative reward. Consider a demonstration $d_j$ of length $L$ that has reward value $r_{d_j}$ computed using \autoref{eqn:total_demo_rob}. Let $s_{bad} \in d_j$ be the states at which a violation of $\varphi$ occurs while $s_{good}$ be the states that do not violate the specification (i.e., $s_{bad} = \{s_j \mid (s_j, a_j) \not\models \psi \}$), then the reward assignment is as shown in \autoref{eqn:bad_reward}.
Intuitively, it penalizes the bad states while ignoring the others since the good states may be part of another demonstration or the learned robot policy that satisfies all requirements.

\vspace*{-10pt}
\mypara{Learner reward} Once the states in each demonstration have been assigned rewards, the next objective is to rank the demonstrations and combine all the rewards from the demonstrations into a cumulative reward that the learner (or robot) will use for finding the desired policy. The demonstrations are sorted by their robustness values to obtain rankings. The learner reward is initialized to zero for all the states in the environment. The resulting reward for the robot is given by $\mathbf{R = \sum_{j=1}^{m} \mathsf{rank(d_j)} \cdot r_{d_j}}$ and then normalized, where $m$ is the number of demonstrations. This equation affects only the states that appear in the demonstrations and the intuition here is that preference is given to higher-ranked demonstrations. By the definition of robustness and its use in reward inferences, it is important to note that ``better" the demonstration, higher the reward. In other words, the rewards are non-decreasing as we move from bad demonstrations to good demonstrations. Hence good demonstrations will strictly have higher reward values and are ranked higher than bad demonstrations. The demonstrations are provided by users on a known map $E_{train}$ and the procedure is formalized in \autoref{alg:reward_inference}.
\begin{wrapfigure}[26]{R}{0.6\textwidth}
    \begin{algorithm}[H]
    \DontPrintSemicolon
    \SetKwInput{Input}{Input}
    \Input{$D$ := set of demonstrations \\ $\Phi$ := set of specifications \\ $E_{train}$ := train map} 
    \KwResult{Infers the learner reward from demonstrations}
    \Begin{
     $r_{d_j} \leftarrow 0$, $\forall d_j \in D$ \; 
     \tcp{Initialize all states to zero}
     Construct $DAG$, compute \autoref{eqn:dag_weights} for $\Phi$ and perform $\mathtt{softmax}$ \;
     \For{$j \leftarrow 1$ to $m$}{
    	\For{$i=1$ to $q$}{
    		Compute $rob_i^j$ \;
    		$r_{d_j} \leftarrow r_{d_j} + w(\varphi_i) \cdot rob_i^j$ \;
    		}
    	\eIf(\tcp*[f]{Good demo}){$rob_i^j > 0, \forall \varphi_i \in \Phi_H$}{
    	 ${r(s_l) \leftarrow \frac{l}{L} \cdot r_{d_j}}$, where $l = 1$ to $L; s_l \in d_j$
    	 }(\tcp*[f]{Bad demo}){Update using \autoref{eqn:bad_reward}}
     }
     Sort all $d \in D$ by their $r_d$ values to estimate $\mathsf{rank}$ \;
     $R \leftarrow \sum_{j=1}^{m} \mathsf{rank(d_j)} \cdot r_{d_j}$ and normalize \;
    }
     \caption{Reward inference from demonstrations. \label{alg:reward_inference}}
    \end{algorithm}
  \end{wrapfigure}

\vspace*{-10pt}
\subsection{Learning Policies from Inferred Rewards}
\label{sec:rl_stl}
\vspace*{-5pt}
In order to learn a policy from the inferred rewards, we can use any of the existing model-free RL algorithms with just 2 modifications to the algorithm during the training step:- (1) reward observation step: during each step of an episode, we record the partial policy of the agent and evaluate it with all the hard specifications $\Phi_H$. The sum of the robustness values of the partial policy for each hard specification is added to the observed reward. This behaves like potential-based reward shaping \citep{NgHR99}, thereby preserving optimality. In the case when a close-to-optimal demonstration is ranked higher than another better demonstration, the algorithm also takes this into account and compensates for the mis-ranking in this step. (2) episode termination step/condition: we terminate the episode when, either the goals are reached or the partial policy violates any hard specification. These two modifications lead to faster and safer learning/exploration. This is especially helpful when agents interact with the environment to learn and the cost of learning unsafe states/behaviors is high (e.g., the robot can get damaged, or may harm humans). In our experiments, we show the effectiveness this approach using standard Q-Learning, which we call \qstl and extend its use for multiple sequential objective MDP. This new \qstl algorithm incorporates RL with verification-in-the-loop method for safer exploration and learning from imperfect demonstrations. The rewards inferred from \autoref{alg:reward_inference}, which we now refer to as feed-forward reward $R_\textit{ff}$ are used to learn the Q-values on a map $E_{test}$ that could be the same as train map or an unseen map of similar size. This $R_\textit{ff}$ is used as a reference/initialization on the new map, hence the requirement that the maps be of similar sizes. We now introduce the notion of feedback reward $R_\textit{fb}$ that the algorithm uses during execution. $R_\textit{fb}$ is initially a copy of $R_\textit{ff}$ and gets updated during each reward observation step of the algorithm as described earlier. Once the Q-values are learned, the algorithm returns a policy from the $start$ state and ending at the desired $goal$ state. We have described a Q-Learning procedure that incorporates STL specifications in learning the Q-values and obtaining a policy, given a start and end state. In order to learn a policy for multiple objectives, consider a set of goal states $Goals = \{g_1, g_2, ..., g_k\}$ where $k$ is the number of objectives or goals. Some specifications can require the robot to achieve the goals in a particular sequential order while others may require the robot to achieve goals without any preference to order. In the case of arbitrary ordering, the number of ways to achieve this is $k!$, hence all the permutations of the goals are stored in a set. For each permutation or ordering\footnote{Partial ordering helps reduce complexity. In the case of particular ordering, this step can be replaced by the desired order and the complexity reduces from $\mathit{k}!$ to $1$.} of the goals $<g_1, g_2, ..., g_{k}>$, a policy is extracted that follows the order: $\pi_p: start \xrightarrow{\text{\qstl}} g_1 \xrightarrow{\text{\qstl}} g_2 \xrightarrow{\text{\qstl}} ... \xrightarrow{\text{\qstl}} g_k$. Each of the final concatenated policies $\pi_p$ is recorded and stored in a dataset represented by $\Pi$. At this stage, the policies in $\Pi$ all satisfy the hard requirements $\Phi_H$ and hence all are valid/feasible trajectories. Finally, the policy that results in maximum robustness w.r.t. the soft requirements $\Phi_S$ is chosen, which imitates the user preferences. The algorithms are detailed in the appendix.


\vspace*{-15pt}
\section{Experiments}
\begin{figure*}[htbp]
\centering
\subfloat[Demo 1]{\includegraphics[width= 1.8in]{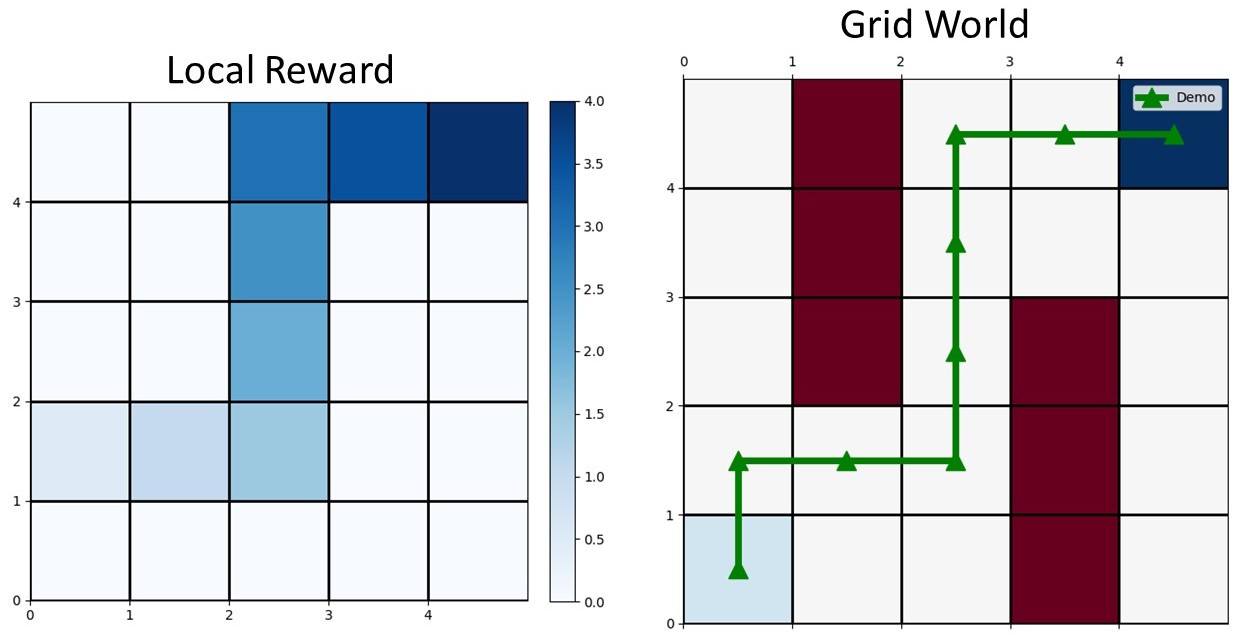}} \hfil
\subfloat[Demo 2]{\includegraphics[width= 1.8in]{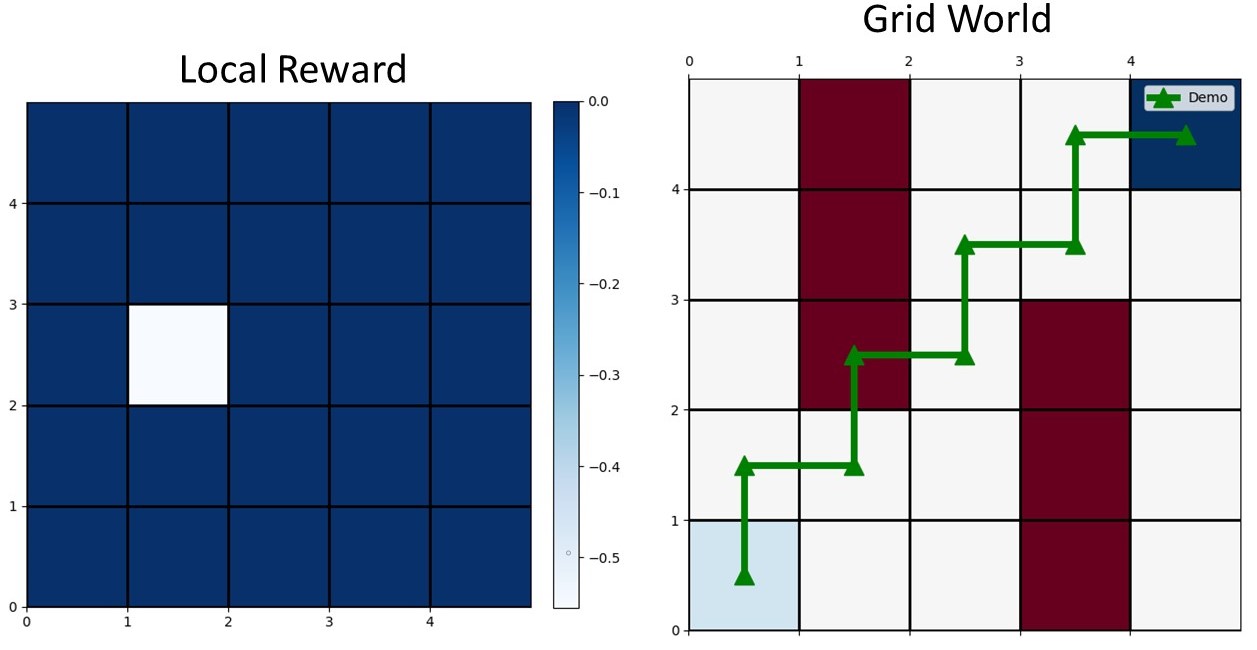}} \hfil
\subfloat[Robot Policy]{\includegraphics[width= 1.8in]{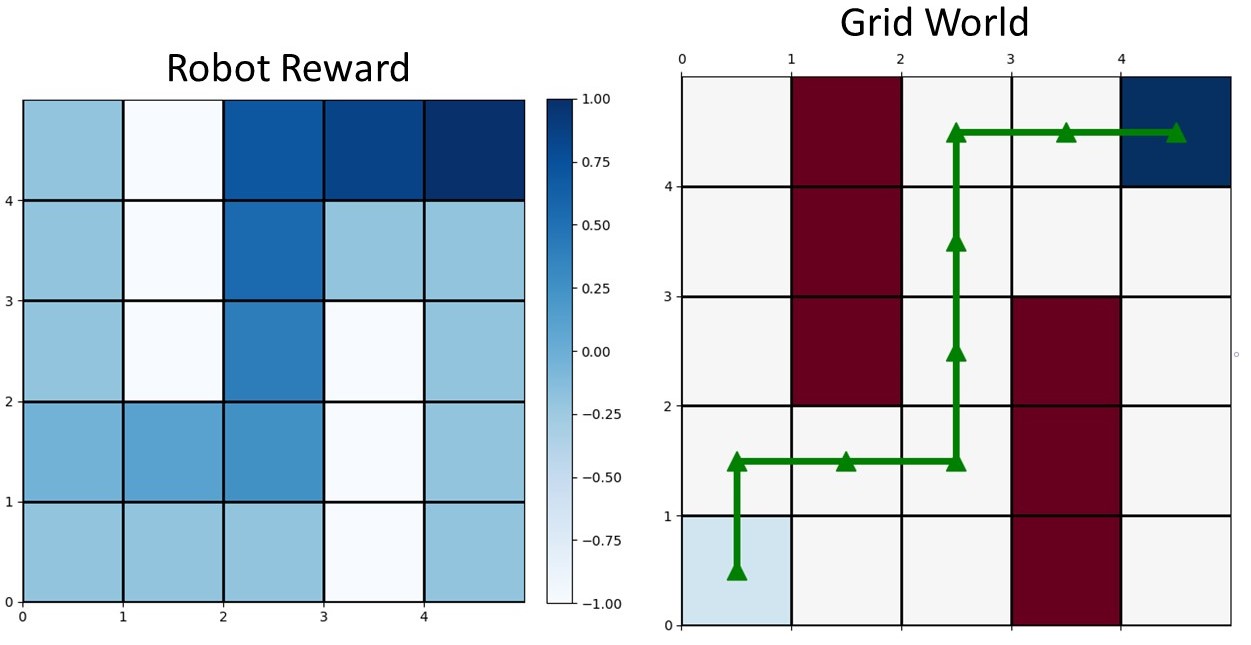}}

\caption{Results: Left figures represent learned rewards. Right figures show the grid-world with start state (light blue), goal (dark blue), obstacles (red) and demonstration/policy (green).}
\label{fig:irl5x5}
\end{figure*}

\vspace*{-15pt}
\mypara{Single-Goal Grid-World} For our experiments, we consider a grid-world environment $E$ consisting of a set of states $S = \{start, goals, obstacles\}$. The map sizes that we used are: $5 \times 5$, $7 \times 7$ and $10 \times 10$; the obstacles were assigned randomly. The distance metric used for this environment is \textit{Manhattan} distance and the STL specifications for this task are:

\begin{enumerate}[topsep=-10pt,itemsep=-1ex,partopsep=1ex,parsep=1ex,leftmargin=1.5em]
	\item Avoid obstacles at all times (hard requirement): $\mathbf{ \varphi_1 := G_{[0,T]} (d_{obs}[t] \ge 1) }$, where $T$ is the length of a demonstration and $d_{obs}$ is the minimum distance of robot from obstacles computed at each step $t$.
	\item Eventually, the robot reaches the goal state (soft requirement): $\mathbf{ \varphi_2 := F_{[0,T]} (d_{goal}[t] < 1) }$, where $d_{goal}$ is the distance of robot from goal computed at each step. $\varphi_2$ depends on $\varphi_1$.
	\item Reach the goal as fast as possible (soft requirement): $\mathbf{ \varphi_3 := F_{[0,T]} (t \le T_{goal}) }$, where $T_{goal}$ is the upper bound of time required to each the goal, which is computed by running breadth-first search algorithm from start to goal state, since the shortest policy must take at least $T_{goal}$ to reach the goal. $\varphi_3$ depends on both $\varphi_1$ and $\varphi_2$.
\end{enumerate}
STL specifications are defined and evaluated using a Matlab toolbox - Breach \citep{breach}. A grid-world point-and-click game was created using $PyGame$ package that showed the locations of start, obstacles and goals. The users provide demonstrations by clicking on their desired states with the task to reach the goal state from start without hitting any obstacles. For this map, we used $m=2$ demonstrations (1 good and 1 bad) from a single user. The demonstrations and resulting robot policy are shown in \autoref{fig:irl5x5}. The blue heatmap figures represent the rewards learned from the demonstrations (darker represent higher rewards). Since hitting a red obstacle is penalized heavily by the hard requirement compared to other states, the rewards in the other safe states and goal state appear similar in value due to the scaling difference. For grid sizes $7 \times 7$ and $10 \times 10$, similar results were observed and each grid had $m=4$ demonstrations (2 good, 1 bad and 1 incomplete). The number of episodes used for training ranged from 3000 to 10000 depending on the complexity (grid size, number and locations of obstacles) of the grid-world. The discount factor $\gamma$ was set to 0.99 and $\epsilon$-greedy strategy for actions was used with $\epsilon = 0.4$. The learning rate used in the experiments is $\alpha = 0.1$.




\vspace*{-15pt}

\vspace*{-10pt}
\mypara{Multi-Goal Grid-World} We also conducted experiments with a grid-world having $k=2$ goals. The specifications used are as follows:
\begin{enumerate}[topsep=0pt,itemsep=-1ex,partopsep=1ex,parsep=1ex,leftmargin=1.5em]
	\item Avoid obstacles at all times (hard requirement): $\mathbf{ \varphi_1 := G_{[0,T]} (d_{obs}[t] \ge 1) }$, where $d_{obs}$ is the minimum distance of robot from obstacles computed at time-step $t$.
	\item Eventually, the robot reaches both goal states in any order (soft requirement): $\mathbf{ \varphi_2 := F_{[0,T]} (d_{goal_1}[t] < 1) \wedge F_{[0,T]} (d_{goal_2}[t] < 1) }$. $\varphi_2$ depends on $\varphi_1$.
	\item Reach the goals as fast as possible (soft requirement): $\mathbf{ \varphi_3 := F_{[0,T]} (t \le T_{goal}) }$, similar to the single-goal grid-world experiment. $\varphi_3$ depends on both $\varphi_1$ and $\varphi_2$.
\end{enumerate}

For the $5 \times 5$ grid, a total of $m=3$ demonstrations were provided (2 good and 1 bad) and for the $7 \times 7$ grid, only $m=2$ good, but sub-optimal demonstrations were provided using similar hyperparameter settings are indicated earlier. Further details are available in the appendix.

\begin{figure}[htbp]
    \centering
    \subfloat[Standard Q-Learning \label{fig:frozenlake8_stats_1}]{\includegraphics[width=1.65in, height=1.4in]{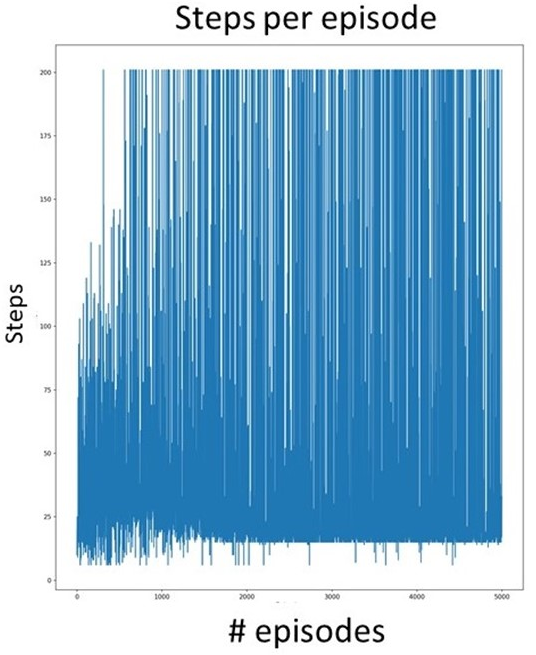}} \hfil
    \subfloat[LfD+STL \label{fig:frozenlake8_stats_2}]{\includegraphics[width=1.65in, height=1.4in]{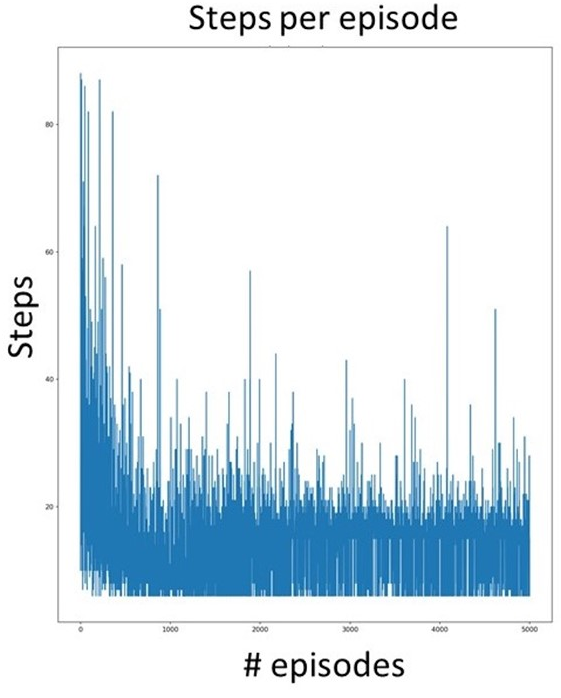}} \hfil
    \subfloat[Mountain Car\label{fig:mountain_car_stats}]{\includegraphics[width=1.65in, height=1.4in]{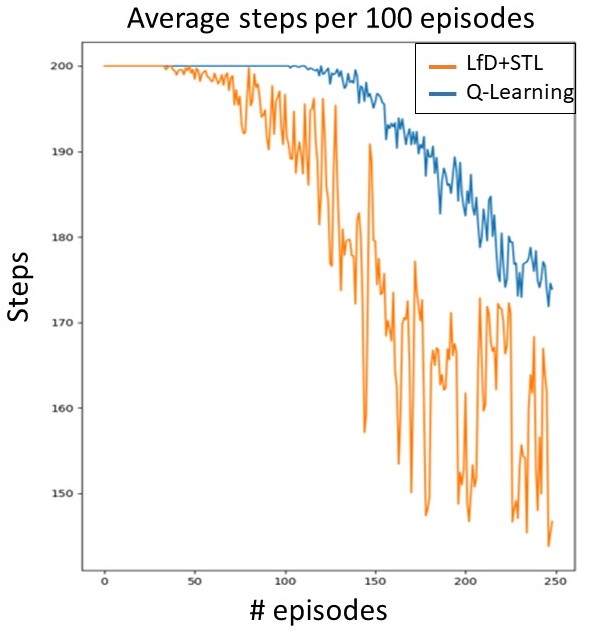}}
    \caption{Comparisons of LfD+STL with hand-crafted rewards+Q-Learning for OpenAI Gym environments. (a) and (b) pertain to Frozenlake and (c) pertains to Mountain Car.} 
\label{fig:openai_stats}
\end{figure}

\vspace*{-10pt}
\mypara{OpenAI Gym} 
The proposed method was tested on the \textit{OpenAI Gym \citep{openai_gym} Frozenlake} environment with both $4 \times 4$ and $8 \times 8$ grid sizes as well as on \textit{Mountain Car}. We compared our method to standard Q-Learning with hand-crafted rewards, based on the number of exploration steps performed by the algorithm in each training episode:- (a) \textit{FrozenLake}: We generated $m=4$ demonstrations by solving the environment using Q-Learning with different hyperparameters to generate different policies. We also modified the $FrozenLake$ grid to relocate the holes, while the goal location remained the same. The specifications used are similar to the single-goal grid-world experiment and are direct representations of the problem statement. Comparisons are shown in Figures \autoref{fig:frozenlake8_stats_1} and \autoref{fig:frozenlake8_stats_2} and we see that our method is able to narrow-down the search exploration space under the same hyperparameter settings. (b) \textit{Mountain Car}: We first abstracted the continuous observation space into $50 \times 50$ grid sizes and generated $m=2$ optimal demonstrations based on a Q-Learning algorithm with preset hyperparameters. We used only one requirement based on the problem definition: $\mathbf{\varphi := \ev_{[0,T]} (d_{flag}[t] \le 0)}$, where $d_{flag}$ is the Manhattan distance between the car and the goal flag positions at time $t$. The comparison with Q-Learning for hand-crafted rewards is summarized in Figure \autoref{fig:mountain_car_stats}. Though there is more variance in the average steps involving our method, we observe that the worst-case average of our algorithm is still better than the best-case average of standard RL. Further details about the demonstration and policy are available in the appendix.

\vspace*{-15pt}
\begin{figure}[htbp]
\centering
\subfloat[Ground Truth \label{fig:gtreward}]{\includegraphics[width=1.5in, height=1.3in]{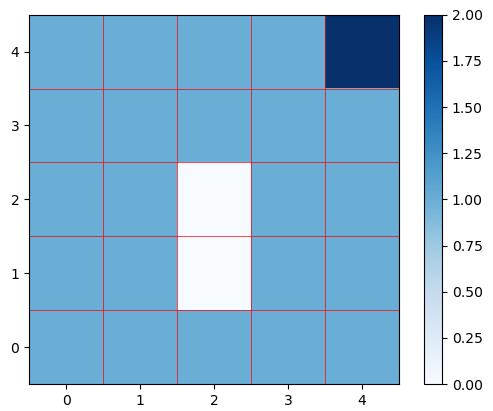}} \hfil
\subfloat[MCE-IRL with 50 optimal demos]{\includegraphics[width=1.5in, height=1.3in]{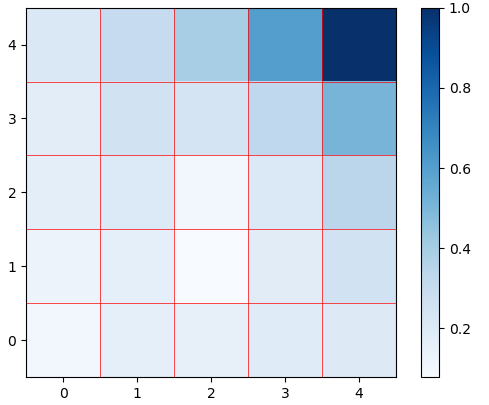}} \hfil
\subfloat[Ours with 3 demos]{\includegraphics[width=1.5in, height=1.3in]{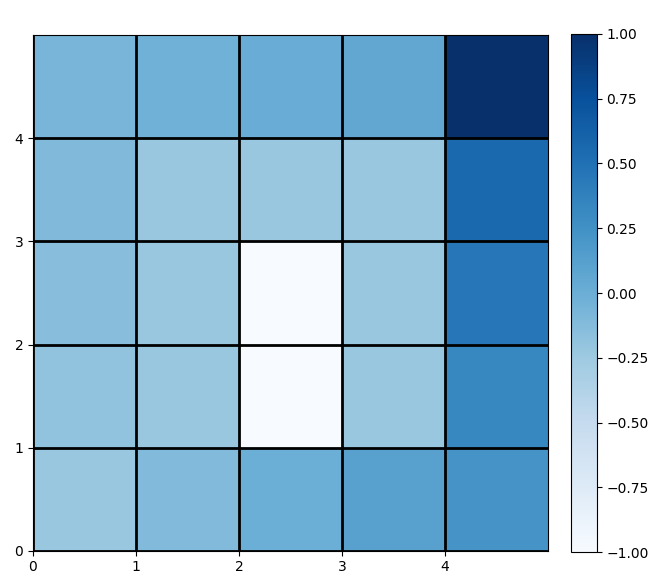}}
\caption{Comparing rewards with ground truth and state-of-the-art MCE-IRL.}
\label{fig:comparisons}
\end{figure}

\vspace*{-10pt} 
\mypara{Discussion and Comparisons} It can be seen that the reward and policy learned by the robot is able to satisfy all the STL requirements from the given initial condition without having the user to explicitly specify/design rewards for the robot and without having to indicate any low-level controls such as robot actions. Because the algorithm automatically performs ranking of demonstrations, it can be interpreted as preference-based learning since it prefers to follow a demonstration that has ``higher" satisfaction of the specifications. Another observation is that our method uses fewer demonstrations and can learn from sub-optimal or imperfect demonstrations. One of the major highlights of our work is that we do not introduce additional hyperparameters and hence any hyperparameter tuning depends on the RL algorithm. We also compared our method with Maximum Causal Entropy IRL (MCE-IRL) \citep{mce_irl_ziebart} on the grid-world and Mountain Car tasks. In the grid-world environment, the ground truth for a $5 \times 5$ grid-world is provided in which the goal is at the top-right corner with reward +2 and the initial state is at the bottom-left. There are 2 states to avoid with reward 0 and every other state where the agent can traverse has a reward of +1 (Figure \autoref{fig:gtreward}). The actual values of the reward are not important since they can be easily interpreted/represented as potential based reward functions which preserve policy optimality. MCE-IRL requires at least 60 $optimal$ demonstrations to recover an approximate reward, whereas our method can recover a more accurate reward with just 3 (2 good and 1 bad) demonstrations (see \autoref{fig:comparisons}). Similar results were obtained with other grid-sizes used in the earlier experiments. For Mountain Car with $50 \times 50$ discretization, both MCE-IRL and our method obtained very similar rewards, with the former requiring at least 10 optimal demonstrations, while the latter used just 2 demonstrations. The ground truth for Mountain Car is provided by the environment itself. Quantitative comparisons are shown in \autoref{table:comparison}. Note that the demonstrations provided for MCE-IRL are all optimal while the demonstrations for our method are mixed (i.e., some good and some bad/sub-optimal). In addition, MCE-IRL does not learn an accurate reward compared to the ground truth. We also noticed that MCE-IRL does not perform well when there are multiple avoid regions/obstacles scattered over the map (e.g., Frozenlake) and in such cases, MCE-IRL requires significantly more demonstrations. On smaller environments, the computation time for inferring rewards is similar for both algorithms. However, as the environment size increases, the computation time and number of demonstrations increase significantly for MCE-IRL. All experiments were conducted on a machine with \textit{AMD Ryzen 7 3700X 8-core} CPU. Lastly, MCE-IRL was not able to recover the reward for the multiple sequential goals, whereas our method was able to do so and found a policy that visited both goals safely and in the shortest time. Unlike many existing IRL techniques, our method also does not involve solving an MDP during the reward inference procedure and the rewards inferred using our method provide better interpretability w.r.t. the specifications. The complexity of the reward inference procedure is polynomial in the length of the specification \citep{stl_complexity} and hence isn't affected by the dimensionality of the state space based on empirical evaluations. As shown in experiments, our algorithm can be used with multiple demonstrators each of whom may be trying to act according to their preferences for the same task. We do not assume uncertainty in sensing and actuation in this setup, and policy synthesis and verification under uncertainty model will be considered as part of future work.

\begin{table}[htbp]
\centering
\begin{tabular}{ |c|c|c|c|c| } 
 \hline
 & \multicolumn{2}{|c|}{\#Demos} &  \multicolumn{2}{|c|}{Avg. Execution Time (in $s$)} \\ \cline{2-5}
 & \textbf{MCE-IRL} & \textbf{Ours} & \textbf{MCE-IRL} & \textbf{Ours} \\ \hline
 ($5 \times 5$) grid & 70 & 3 & 3.77 & 2.62 \\ \hline 
 ($7 \times 7$) grid & 150 & 5 & 6.81 & 2.74 \\ \cline{1-5} 
 FrozenLake-4 & 150 & 4 & 3.96 & 2.81 \\ \hline
 FrozenLake-8 & 800 & 5 & 13.18 & 3.11 \\ \hline
 Mt. Car & 10-20 & 2-3 & $>60$ & 2.95 \\
 \hline
\end{tabular}
\caption{Quantitative comparisons between MCE-IRL and our method for different environments.}
\label{table:comparison}
\end{table}

        
        

\vspace*{-20pt}
\vspace*{-5pt}

\vspace*{-12pt}
\section{Related Work and Conclusion}
\vspace*{-15pt}
\mypara{Related Work} The use of formal methods with LfD has been explored by the authors of \citep{belta_lfd} who proposed to learn tasks with complex structures by combining temporal logics with reinforcement LfD. It involves designing a special logic, augmenting a finite-state automaton with MDP and then using behavioral cloning for policy initialization and policy gradient to train the agent, but relies on optimal/perfect demonstrations. The final learned policy is not found to be very robust w.r.t. the specifications. The authors in \citep{wenchao_cav} propose a counterexample-guided approach using probabilistic computation tree logics for safety-aware AL. Similar to our work, they perform a verification-in-the-loop approach by embedding the logic-checking mechanism inside the training loop and automatically generating a counterexample in case of violation. Most recently, the authors of \citep{mvc_specs} seek to learn task specifications from expert demonstrations using the principle of causal entropy and model-based MDPs. A similar work \citep{KasenbergS17} aims to infer linear temporal logic (LTL) specifications from agent behavior in MDPs as a path to interpretable AL. There have also been several works that utilize the robustness semantics of STL to describe reward functions in the RL domain \citep{aksaray_q-learning_2016, li_reinforcement_2017}. In \citep{anandb_rl}, the authors propose to use STL and its quantitative semantics to generate locally shaped reward functions, that considers the robustness of a system trajectory for some finite window of its execution resulting in a local approximation of the direction of the system trajectory. Traditional motion-planning with chance constraints has been investigated \citep{ono_plan} to produce plans with bounded risk. However, this work involves solving numerous constraints, manually ranking or selecting among various feasible paths and expert-designed costs. Our work differs in that the constraints are now replaced by formal specifications and the costs are rewards that are inferred. Based on these, the demonstrations are automatically ranked and a new robot policy is learned. Existing works that learn from suboptimal/imperfect demonstrations do so by filtering such demonstrations or classifying suboptimal demonstrations when most of the other demonstrations are optimal \citep{ravichandar_recent_2020}. \\
\vspace*{-18pt}

\mypara{Conclusion} We introduced a framework that combines human demonstrations and high-level STL specifications to: (1) quantitatively evaluate and rank demonstrations and (2) infer \textit{non-Markovian} rewards for a robot such that the computed policy is able to satisfy all specifications. We conducted several discrete-world experiments to justify the effectiveness of our method. This approach would provide new directions for safety and interpretability of robot control policies and verification of model-free learning methods. Since our framework (a) does not introduce additional hyperparameters, (b) can learn from a few demonstrations and (c) facilitates safer and faster learning, it is appropriate for non-expert users and real-world applications. It is also well suited for applications where the maps are known beforehand but there exist dynamic obstacles in the map, such as for robots in household and warehouse environments, space exploration rovers, etc.

	


\clearpage
\acknowledgments{The authors thank the reviewers for their time and insightful feedback. The authors also gratefully acknowledge the support of the National Science Foundation under the grant FMitF grant CCF-1837131, CPS grant CNS-1932620, and the support from Toyota Motors North America R\&D.}


\bibliography{myrefs, cps}  

\begin{thebibliography}{31}
\providecommand{\natexlab}[1]{#1}
\providecommand{\url}[1]{\texttt{#1}}
\expandafter\ifx\csname urlstyle\endcsname\relax
  \providecommand{\doi}[1]{doi: #1}\else
  \providecommand{\doi}{doi: \begingroup \urlstyle{rm}\Url}\fi

\bibitem[Atkeson and Schaal(1997)]{lfd_schaal1}
C.~G. Atkeson and S.~Schaal.
\newblock Robot learning from demonstration.
\newblock In D.~H. Fisher, editor, \emph{Proceedings of the Fourteenth
  International Conference on Machine Learning {(ICML} 1997), Nashville,
  Tennessee, USA, July 8-12, 1997}, pages 12--20. Morgan Kaufmann, 1997.

\bibitem[Schaal(1996)]{lfd_schaal2}
S.~Schaal.
\newblock Learning from demonstration.
\newblock In M.~Mozer, M.~I. Jordan, and T.~Petsche, editors, \emph{Advances in
  Neural Information Processing Systems 9, NIPS, Denver, CO, USA, December 2-5,
  1996}, pages 1040--1046. {MIT} Press, 1996.

\bibitem[Abbeel and Ng(2004)]{abbeel_ng}
P.~Abbeel and A.~Y. Ng.
\newblock Apprenticeship learning via inverse reinforcement learning.
\newblock In \emph{Proceedings of the Twenty-First International Conference on
  Machine Learning}, ICML ’04, page~1, New York, NY, USA, 2004. Association
  for Computing Machinery.

\bibitem[Ng and Russell(2000)]{Ng_russell}
A.~Y. Ng and S.~J. Russell.
\newblock Algorithms for inverse reinforcement learning.
\newblock In P.~Langley, editor, \emph{Proceedings of the Seventeenth
  International Conference on Machine Learning {(ICML} 2000), Stanford
  University, Stanford, CA, USA, June 29 - July 2, 2000}, pages 663--670.
  Morgan Kaufmann, 2000.

\bibitem[Ziebart et~al.(2008)Ziebart, Maas, Bagnell, and
  Dey]{ziebart_maximumentropy}
B.~D. Ziebart, A.~L. Maas, J.~A. Bagnell, and A.~K. Dey.
\newblock Maximum entropy inverse reinforcement learning.
\newblock In \emph{Aaai}, volume~8, pages 1433--1438. Chicago, IL, USA, 2008.

\bibitem[Torabi et~al.(2018)Torabi, Warnell, and Stone]{bco_stone}
F.~Torabi, G.~Warnell, and P.~Stone.
\newblock Behavioral cloning from observation.
\newblock In J.~Lang, editor, \emph{Proceedings of the Twenty-Seventh
  International Joint Conference on Artificial Intelligence, {IJCAI} 2018, July
  13-19, 2018, Stockholm, Sweden}, pages 4950--4957. ijcai.org, 2018.

\bibitem[Ravichandar et~al.(2020)Ravichandar, Polydoros, Chernova, and
  Billard]{ravichandar_recent_2020}
H.~Ravichandar, A.~S. Polydoros, S.~Chernova, and A.~Billard.
\newblock Recent {Advances} in {Robot} {Learning} from {Demonstration}.
\newblock \emph{Annual Review of Control, Robotics, and Autonomous Systems},
  3\penalty0 (1):\penalty0 297--330, May 2020.
\newblock ISSN 2573-5144, 2573-5144.

\bibitem[Hussein et~al.(2017)Hussein, Gaber, Elyan, and
  Jayne]{hussein_imitation_2017}
A.~Hussein, M.~M. Gaber, E.~Elyan, and C.~Jayne.
\newblock Imitation {Learning}: {A} {Survey} of {Learning} {Methods}, Apr.
  2017.

\bibitem[Osa et~al.(2018)Osa, Pajarinen, Neumann, Bagnell, Abbeel, and
  Peters]{osa_algorithmic_2018}
T.~Osa, J.~Pajarinen, G.~Neumann, J.~A. Bagnell, P.~Abbeel, and J.~Peters.
\newblock An {Algorithmic} {Perspective} on {Imitation} {Learning}.
\newblock \emph{Foundations and Trends in Robotics}, 7\penalty0 (1-2):\penalty0
  1--179, 2018.
\newblock ISSN 1935-8253, 1935-8261.
\newblock arXiv: 1811.06711.

\bibitem[Donz\'e et~al.(2015)Donz\'e, Jin, Deshmukh, and
  Seshia]{donze_automotive_2015}
A.~Donz\'e, X.~Jin, J.~V. Deshmukh, and S.~A. Seshia.
\newblock Automotive systems requirement mining using breach.
\newblock In \emph{2015 {{American Control Conference}} ({{ACC}})}, pages
  4097--4097, July 2015.

\bibitem[Kapinski et~al.(2016)Kapinski, Jin, Deshmukh, Donze, Yamaguchi, Ito,
  Kaga, Kobuna, and Seshia]{kapinski_st-lib_2016}
J.~Kapinski, X.~Jin, J.~Deshmukh, A.~Donze, T.~Yamaguchi, H.~Ito, T.~Kaga,
  S.~Kobuna, and S.~Seshia.
\newblock {{ST}}-{{Lib}}: {{A Library}} for {{Specifying}} and {{Classifying
  Model Behaviors}}.
\newblock {{SAE Technical Paper}} 2016-01-0621, {SAE International},
  Warrendale, PA, Apr. 2016.

\bibitem[Bartocci et~al.(2018)Bartocci, Deshmukh, Donz\'e, Fainekos, Maler,
  N{\v i}ckovi\'c, and Sankaranarayanan]{bartocci_specification-based_2018}
E.~Bartocci, J.~Deshmukh, A.~Donz\'e, G.~Fainekos, O.~Maler, D.~N{\v
  i}ckovi\'c, and S.~Sankaranarayanan.
\newblock Specification-{{Based Monitoring}} of {{Cyber}}-{{Physical Systems}}:
  {{A Survey}} on {{Theory}}, {{Tools}} and {{Applications}}.
\newblock In E.~Bartocci and Y.~Falcone, editors, \emph{Lectures on {{Runtime
  Verification}}: {{Introductory}} and {{Advanced Topics}}}, pages 135--175.
  {Springer International Publishing}, Cham, 2018.

\bibitem[Deshmukh et~al.(2017)Deshmukh, Donz\'e, Ghosh, Jin, Juniwal, and
  Seshia]{deshmukh_robust_2017}
J.~V. Deshmukh, A.~Donz\'e, S.~Ghosh, X.~Jin, G.~Juniwal, and S.~A. Seshia.
\newblock Robust online monitoring of signal temporal logic.
\newblock \emph{Formal Methods in System Design}, 51\penalty0 (1):\penalty0
  5--30, Aug. 2017.
\newblock ISSN 1572-8102.

\bibitem[Fainekos and Pappas(2009)]{fainekos_robustness_2009}
G.~E. Fainekos and G.~J. Pappas.
\newblock Robustness of temporal logic specifications for continuous-time
  signals.
\newblock \emph{Theoretical Computer Science}, 410\penalty0 (42):\penalty0
  4262--4291, 2009.

\bibitem[Sutton and Barto(2018)]{sutton_reinforcement_2018}
R.~S. Sutton and A.~G. Barto.
\newblock \emph{Reinforcement Learning: An Introduction}.
\newblock Adaptive Computation and Machine Learning Series. {The MIT Press},
  Cambridge, MA, second edition, 2018.

\bibitem[Leike et~al.(2017)Leike, Martic, Krakovna, Ortega, Everitt, Lefrancq,
  Orseau, and Legg]{leike_ai_2017}
J.~Leike, M.~Martic, V.~Krakovna, P.~A. Ortega, T.~Everitt, A.~Lefrancq,
  L.~Orseau, and S.~Legg.
\newblock Ai safety gridworlds, 2017.

\bibitem[Ng et~al.(1999)Ng, Harada, and Russell]{NgHR99}
A.~Y. Ng, D.~Harada, and S.~J. Russell.
\newblock Policy invariance under reward transformations: Theory and
  application to reward shaping.
\newblock In I.~Bratko and S.~Dzeroski, editors, \emph{Proceedings of the
  Sixteenth International Conference on Machine Learning {(ICML} 1999), Bled,
  Slovenia, June 27 - 30, 1999}, 1999.

\bibitem[Donz{\'{e}}(2010)]{breach}
A.~Donz{\'{e}}.
\newblock Breach, {A} toolbox for verification and parameter synthesis of
  hybrid systems.
\newblock In T.~Touili, B.~Cook, and P.~B. Jackson, editors, \emph{Computer
  Aided Verification, 22nd International Conference, {CAV} 2010, Edinburgh, UK,
  July 15-19, 2010. Proceedings}, volume 6174 of \emph{Lecture Notes in
  Computer Science}, pages 167--170. Springer, 2010.

\bibitem[Brockman et~al.(2016)Brockman, Cheung, Pettersson, Schneider,
  Schulman, Tang, and Zaremba]{openai_gym}
G.~Brockman, V.~Cheung, L.~Pettersson, J.~Schneider, J.~Schulman, J.~Tang, and
  W.~Zaremba.
\newblock Openai gym.
\newblock \emph{CoRR}, abs/1606.01540, 2016.

\bibitem[Ziebart(2010)]{mce_irl_ziebart}
B.~D. Ziebart.
\newblock \emph{Modeling Purposeful Adaptive Behavior with the Principle of
  Maximum Causal Entropy}.
\newblock PhD thesis, Carnegie Mellon University, USA, 2010.

\bibitem[Maler and Nickovic(2004)]{stl_complexity}
O.~Maler and D.~Nickovic.
\newblock Monitoring temporal properties of continuous signals.
\newblock In Y.~Lakhnech and S.~Yovine, editors, \emph{Formal Techniques,
  Modelling and Analysis of Timed and Fault-Tolerant Systems}, pages 152--166,
  Berlin, Heidelberg, 2004. Springer Berlin Heidelberg.
\newblock ISBN 978-3-540-30206-3.

\bibitem[Li et~al.(2018)Li, Ma, and Belta]{belta_lfd}
X.~Li, Y.~Ma, and C.~Belta.
\newblock Automata guided reinforcement learning with demonstrations.
\newblock \emph{CoRR}, abs/1809.06305, 2018.

\bibitem[Zhou and Li(2018)]{wenchao_cav}
W.~Zhou and W.~Li.
\newblock Safety-aware apprenticeship learning.
\newblock In H.~Chockler and G.~Weissenbacher, editors, \emph{Computer Aided
  Verification - 30th International Conference, {CAV} 2018, UK, July 14-17,
  2018, Proceedings, Part {I}}, volume 10981 of \emph{Lecture Notes in Computer
  Science}, pages 662--680. Springer, 2018.

\bibitem[Vazquez{-}Chanlatte and Seshia(2020)]{mvc_specs}
M.~Vazquez{-}Chanlatte and S.~A. Seshia.
\newblock Maximum causal entropy specification inference from demonstrations.
\newblock In S.~K. Lahiri and C.~Wang, editors, \emph{Computer Aided
  Verification - 32nd International Conference, {CAV} 2020, Los Angeles, CA,
  USA, July 21-24, 2020, Proceedings, Part {II}}, volume 12225 of \emph{Lecture
  Notes in Computer Science}, pages 255--278. Springer, 2020.

\bibitem[Kasenberg and Scheutz(2017)]{KasenbergS17}
D.~Kasenberg and M.~Scheutz.
\newblock Interpretable apprenticeship learning with temporal logic
  specifications.
\newblock In \emph{56th {IEEE} Annual Conference on Decision and Control, {CDC}
  2017, Melbourne, Australia, December 12-15, 2017}, pages 4914--4921. {IEEE},
  2017.

\bibitem[Aksaray et~al.(2016)Aksaray, Jones, Kong, Schwager, and
  Belta]{aksaray_q-learning_2016}
D.~Aksaray, A.~Jones, Z.~Kong, M.~Schwager, and C.~Belta.
\newblock Q-{{Learning}} for robust satisfaction of signal temporal logic
  specifications.
\newblock In \emph{2016 {{IEEE}} 55th {{Conference}} on {{Decision}} and
  {{Control}} ({{CDC}})}, pages 6565--6570, Dec. 2016.

\bibitem[Li et~al.(2017)Li, Vasile, and Belta]{li_reinforcement_2017}
X.~Li, C.~Vasile, and C.~Belta.
\newblock Reinforcement learning with temporal logic rewards.
\newblock In \emph{2017 {{IEEE}}/{{RSJ International Conference}} on
  {{Intelligent Robots}} and {{Systems}} ({{IROS}})}, pages 3834--3839, Sept.
  2017.

\bibitem[Balakrishnan and Deshmukh(2019)]{anandb_rl}
A.~Balakrishnan and J.~V. Deshmukh.
\newblock Structured reward shaping using signal temporal logic specifications.
\newblock In \emph{2019 {IEEE/RSJ} International Conference on Intelligent
  Robots and Systems, {IROS} 2019, Macau, SAR, China, November 3-8, 2019},
  pages 3481--3486. {IEEE}, 2019.

\bibitem[Ono et~al.(2013)Ono, Williams, and Blackmore]{ono_plan}
M.~Ono, B.~C. Williams, and L.~Blackmore.
\newblock Probabilistic planning for continuous dynamic systems under bounded
  risk.
\newblock \emph{J. Artif. Intell. Res.}, 46:\penalty0 511--577, 2013.

\bibitem[Jak{\v s}i\'c et~al.(2018)Jak{\v s}i\'c, Bartocci, Grosu, Nguyen, and
  Ni{\v c}kovi\'c]{jaksic_quantitative_2018}
S.~Jak{\v s}i\'c, E.~Bartocci, R.~Grosu, T.~Nguyen, and D.~Ni{\v c}kovi\'c.
\newblock Quantitative monitoring of {{STL}} with edit distance.
\newblock \emph{Formal Methods in System Design}, 53\penalty0 (1):\penalty0
  83--112, Aug. 2018.
\newblock ISSN 1572-8102.

\bibitem[Donz\'e and Maler(2010)]{donze_robust_2010}
A.~Donz\'e and O.~Maler.
\newblock Robust satisfaction of temporal logic over real-valued signals.
\newblock In \emph{International {{Conference}} on {{Formal Modeling}} and
  {{Analysis}} of {{Timed Systems}}}, pages 92--106. {Springer}, 2010.

\end{thebibliography}

\clearpage
\section*{Appendix}
\appendix
\section{Reinforcement Learning (RL)}
\vspace*{-10pt}
\begin{definition}[Model-Free Markov Decision Process (MDP)]
    It is a tuple $M = \left(S, A, R, \gamma\right)$ where
    \begin{itemize}[topsep=0pt,itemsep=-1ex,partopsep=1ex,parsep=1ex,leftmargin=1.5em]
        \item $S$ is the state space of the system;
        \item $A$ is the set of actions that can be performed on the system;
        \item $R$ is a reward function that typically maps either some $s \in S$
        or some transition $\delta \in S \times A \times S$ to $\Reals$;
        \item $\gamma$ is the discount factor for the MDP.
    \end{itemize}
\end{definition}
\vspace*{-10pt}
\section{Quantitative Semantics of STL}
\vspace*{-10pt}
A basic example of STL and the mathematical definition of quantitative semantics are described below.
\begin{example}
    Consider the signal $x(t)$ obtained by sampling the function
    $\sin(2\pi t)$ at times $t_0,t_1,\ldots$, where $t_j = j\times
    0.125$ (shown in \autoref{fig:example1}).  Consider the
    formula $\alw(x(t) \ge -1)$, which requires that starting at time
    $0$, $x(t)$ is always greater than $-1$ (at each sample point).
    Consider the formula $\ev_{[0,3]}(\alw_{[0,1]}(x(t)\ge0))$. This formula
    requires that there is some time (say $\tau$) such that between
    times $[\tau,\tau+1]$, $x(\tau)$ is always greater than $0$.
    Considering that $x(t)$ is a sampling of a sinusoid with period
    $1$, this formula is also satisfied by $x(t)$.
\end{example}
In addition to the Boolean satisfaction semantics for STL, various researchers
have proposed quantitative semantics for STL, \citep{fainekos_robustness_2009, jaksic_quantitative_2018} that compute the degree of satisfaction (or \textit{robust satisfaction values})
of STL properties by traces generated by a system.
\begin{definition}[Quantitative Semantics for Signal Temporal Logic]%
\label{def:quantitative}
    Given an algebraic structure $(\oplus,\otimes,\top,\bot)$, we define the
    quantitative semantics for an arbirtary signal $\sig$ against an STL formula
    $\varphi$ at time $t$ as follows:
    \begin{center}
            \begin{minipage}{0.8\textwidth}
    \begin{tabular*}{\linewidth}{@{\extracolsep{\fill} } r@{}l}
      $\varphi$ &  $\robustness{\varphi}{t}$ \\
      \hline
      $\mathit{true}$/$\mathit{false}$ & $\top$/$\bot$ \\
      $\mu$                       & $f(\sig(t))$ \\
      $\neg \varphi$              & $-\robustness{\varphi}{t}$ \\
      
      $\varphi_1 \wedge \varphi_2$ &
      $\otimes(\robustness{\varphi_1}{t},\robustness{\varphi_2}{t})$ \\
      
      $\varphi_1 \vee \varphi_2$ & 
      $\oplus(\robustness{\varphi_1}{t},\robustness{\varphi_2}{t})$ \\
      
      $\alw_I(\varphi)$ &  $\otimes_{\tau\in t+I}(\robustness{\varphi}{\tau})$ \\
      
      $\ev_I(\varphi)$ &  $\oplus_{\tau\in t+I}(\robustness{\varphi}{\tau})$ \\

      $\varphi \until_I \psi$ & $\oplus_{\tau_1\in t+I}
      (\otimes(\robustness{\psi}{\tau_1},\otimes_{\tau_2\in[t,\tau_1)}(\robustness{\varphi}{\tau_2}))$
      
\end{tabular*}
\end{minipage}
\end{center}
\end{definition}

The above definition is quite abstract: it does not give specific
interpretations to the elements $\top,\bot$ or the operators $\otimes$ and
$\oplus$. In the original definitions of robust satisfaction proposed in
\citep{fainekos_robustness_2009,donze_robust_2010}, the interpretation was to set
$\top = +\infty$, $\bot = -\infty$, and $\oplus = \max$, and $\otimes = \min$.
\vspace*{-10pt}
\section{Algorithms}
\vspace*{-10pt}
\subsection{Q-Learning with STL}
\vspace*{-5pt}
This algorithm is a modification of the standard Q-Learning algorithm that integrates STL specifications during training as described in the main paper (\autoref{alg:ql_stl}).


\begin{algorithm}
\DontPrintSemicolon
\SetKwInput{Input}{Input}
\Input{$R_{fb}$ := feedback reward \\ $\Phi$ := set of specifications \\ $E_{test}$ := test map \\ $start$ := start state \\ $goal$ := goal state} 
\KwResult{Computes agent policy from $start$ to $goal$}
\Begin{
 $Q[s, a] \leftarrow 0, \forall s \in S, \forall a \in A$ \tcp*[r]{Initialize}
 \For{$ep \leftarrow 1$ to $\# Episodes$}{
    $\pi_{partial} \leftarrow \emptyset$ \;
	$s \leftarrow start$ \;
	$\pi_{partial} \leftarrow \pi_{partial} +  s$ \tcp*[r]{the `+' operator here represents concatenation or append.}
	$done \leftarrow False$ \;
	\While{not $done$}{
	$a \leftarrow \epsilon$-greedy strategy \;
	Observe next state $s'$ and reward $r$ based on action $a$ \;
	$\pi_{partial} \leftarrow \pi_{partial} +  s'$ \tcp*[r]{the `+' operator here represents concatenation or append.}
	$r \leftarrow r + \sum_{\varphi \in \Phi_{H}} \rho(\varphi, \pi_{partial}, t)$ \tcp*[r]{adding robustness w.r.t. $\Phi_H$ to reward as feedback}
	$Q[s, a] \leftarrow Q[s, a] + \alpha[r + \gamma max_a Q[s', a] - Q[s, a]]$ \;
	\If(\tcp*[f]{if goal state reached or policy violates any $\Phi_{H}$}){$(s' \in Goals)$ or ($\rho(\varphi, \pi_{partial}, t) < 0$, for any $\varphi \in \Phi_H)$}{$done \leftarrow True$}
	}
 }
 $\pi \leftarrow$ policy from $start$ \;
 \Return $\pi$ \;
 }
 
\caption{$Q_{stl}$ algorithm\label{alg:ql_stl}}
\end{algorithm}
\vspace*{-10pt}
\subsection{Learning multi-objective robot policy from inferred rewards}
\vspace*{-5pt}
We formalize the algorithm that utilizes \qstl to obtain a policy between two states of the environment and then concatenates the piece-wise policies to form a final control policy for the robot that is able to visit all the goals/objectives as per the task specification (\autoref{alg:robot_policy}).


\begin{algorithm}
\DontPrintSemicolon
\SetKwInput{Input}{Input}
\Input{$R_{ff}$ := Inferred rewards \\ $\Phi$ := set of specifications \\ $E_{test}$ := test map} 
\KwResult{Learns multi-objective robot policy from inferred rewards}
\Begin{
 $P \leftarrow PermutationSet(Goals)$ \tcp*[r]{generates permutation of all goal or objective states.}
 $R_{fb} \leftarrow R_{ff}$ \;
 $\Pi \leftarrow \emptyset$ \;
 \For(\tcp*[f]{$p$ is an ordering of goals $<g_1, g_2, ..., g_{k}>$}){$p \in P$}{
  $\pi_p \leftarrow Q_{stl}(R_{fb}, \Phi, E_{test}, Init, g_1)$ \tcp*[r]{from \autoref{alg:ql_stl}} \;
  \For {$i \leftarrow 1$ to $|Goals|-1$}{
  $\pi_p \leftarrow \pi_p + Q_{stl}(R_{fb}, \Phi, E_{test}, g_i, g_{i+1})$ \tcp*[r]{from \autoref{alg:ql_stl}} \;
  }
   $\Pi \leftarrow \Pi \cup \pi_p$ \;
 }
 \tcp{The resulting policies satisfy all hard requirements}
 $\pi^* \leftarrow \argmax_{\pi \in \Pi} \sum_{\varphi \in \Phi_{S}} \rho(\varphi, \pi, t)$ \tcp*[r]{Policy that maximizes robustness of all soft-requirements}
 }
 
\caption{Learning multi-objective robot policy from inferred rewards\label{alg:robot_policy}}
\end{algorithm}
\vspace*{-10pt}
\section{Experiments}
\vspace*{-10pt}
\subsection{PyGame Setting}
\begin{figure}
\centering
\begin{minipage}{.5\textwidth}
  \centering
  \includegraphics[scale=0.4]{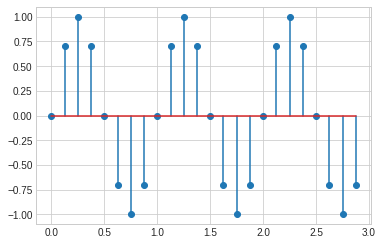}
  \captionof{figure}{Properties on a $\sin$ wave.\label{fig:example1}}
\end{minipage}%
\begin{minipage}{.5\textwidth}
  \centering
     \subfloat[Grid-world game setup]{\includegraphics[scale=0.7]{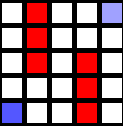}} \hfil
    \subfloat[Example user demonstration (in green)]{\includegraphics[scale=0.7]{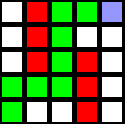}}
  \captionof{figure}{$PyGame$ user-interface. \label{fig:game}}
\end{minipage}
\end{figure}

\vspace*{-5pt}
An screenshot of the grid-world created using $PyGame$ package is shown in \autoref{fig:game} along with a sample demonstration. It is a point-and-click game/interface for a user to provide demonstrations. The task is to select or click on cells starting from the dark blue cell (bottom-left) and ending in the light blue cell (top-right). The red cells represent ``avoid" regions or obstacles.

\begin{figure*}[htbp]
\centering
\subfloat[Demo 1 (good)]{\includegraphics[width= 1.8in]{final_grid5/demo1}} \hfil
\subfloat[Demo 2 (bad)]{\includegraphics[width= 1.8in]{final_grid5/demo2}} \hfil
\subfloat[Robot Policy]{\includegraphics[width= 1.8in]{final_grid5/robot}}

\subfloat[Ground Truth Reward]{\includegraphics[scale = 0.25]{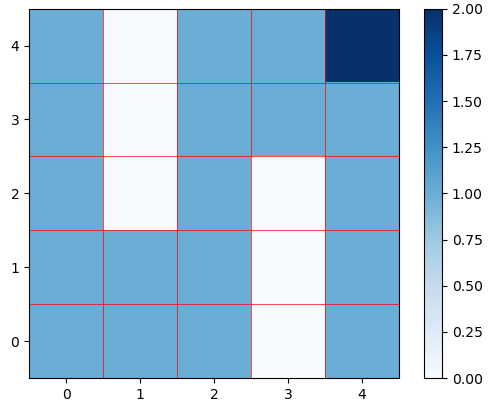}} \hfil
\subfloat[MCE-IRL Reward]{\includegraphics[scale = 0.25]{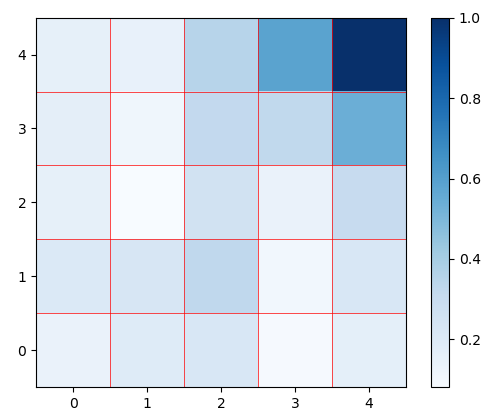}}
\caption{Results for $5 \times 5$ grid-world.}
\label{fig:irl_5x5_latest}
\end{figure*}

To illustrate with an example, consider the $5 \times 5$ grid-world for single goal as shown in \autoref{fig:irl_5x5_latest} and described in the main article. Two demonstrations are provided (1 good and 1 bad). In the good demonstration, the reward is assigned to every state appearing in the demonstration while other states are kept at zero. The rewards increase from start state to the goal so as to guide the robot towards the goal. In the bad demonstration, one of the states coincides with an obstacle and only that state is penalized. The final robot reward is a linear combination of the demonstration rewards. We also show the ground truth reward for this map and the rewards extracted using MCE-IRL with over 40 optimal demonstrations. It is clear again that our method infers more interpretable rewards than the state of the art MCE-IRL, while using far fewer demonstrations. We obtain similar comparisons for other grids.
\vspace*{-15pt}
\vspace*{-15pt}
\subsection{Multiple Sequential Goal Grid-World}
\begin{figure*}[htbp]
\centering
\subfloat[Demo 1]{\includegraphics[width= 1.8in]{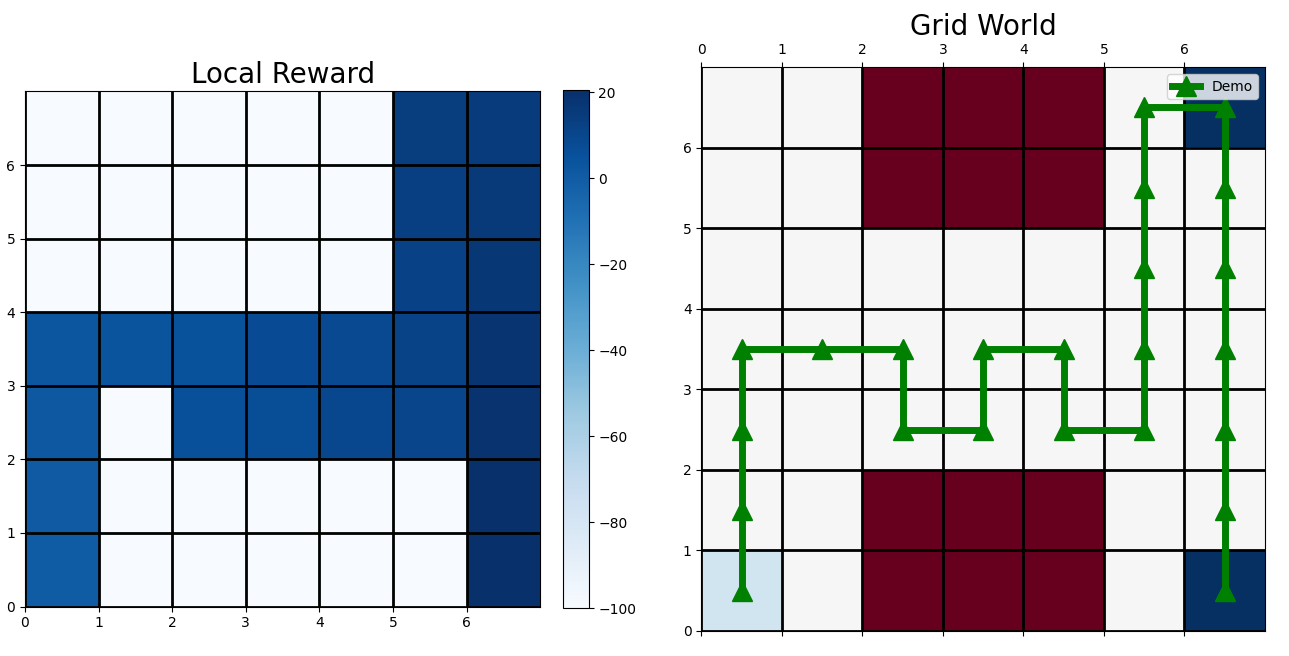}} \hfil
\subfloat[Demo 2]{\includegraphics[width= 1.8in]{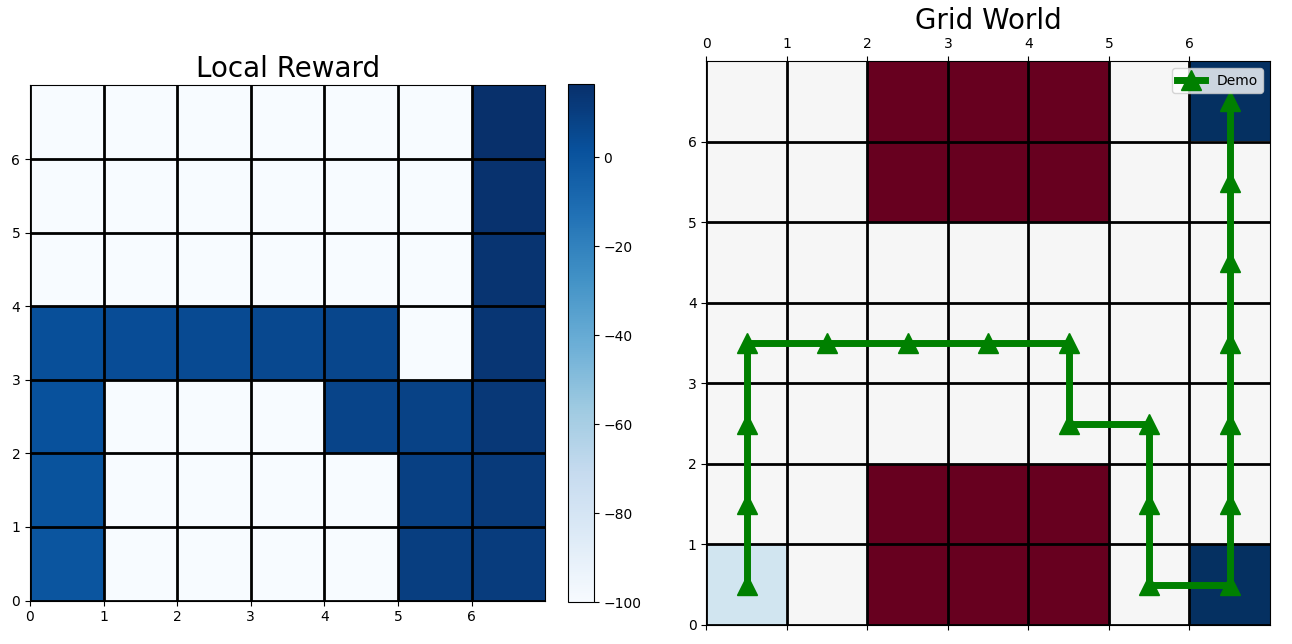}} \hfil
\subfloat[Robot Policy]{\includegraphics[width= 1.8in]{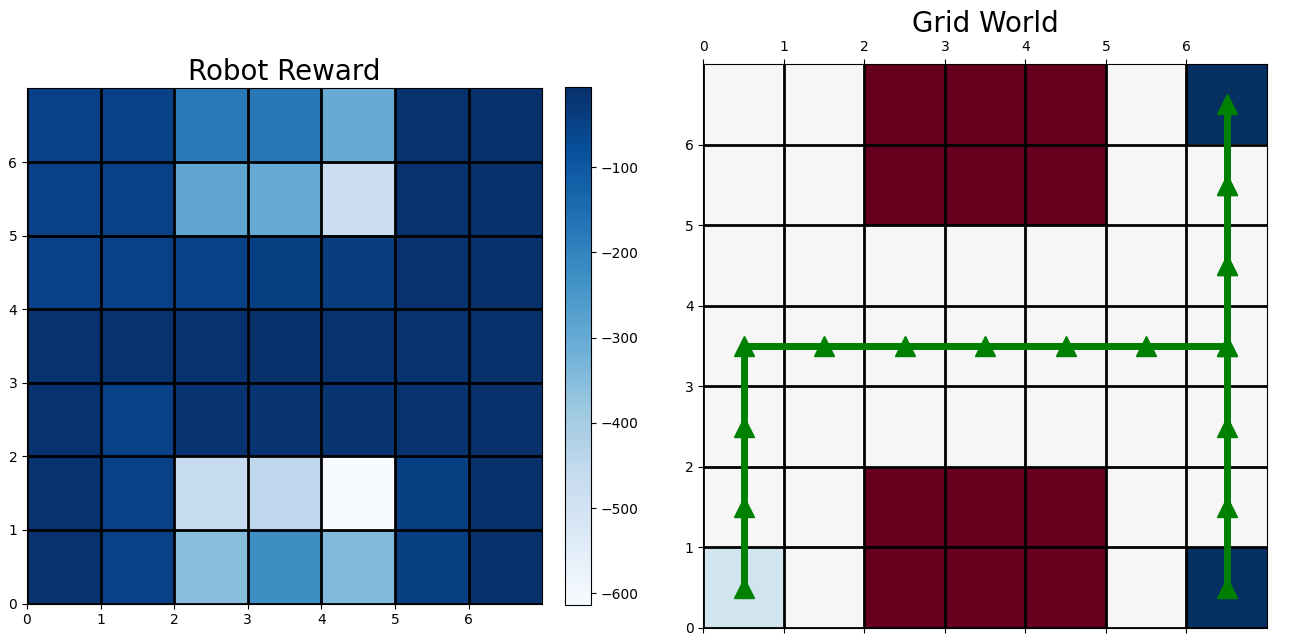}} 

\subfloat[Ground Truth Reward]{\includegraphics[scale = 0.25]{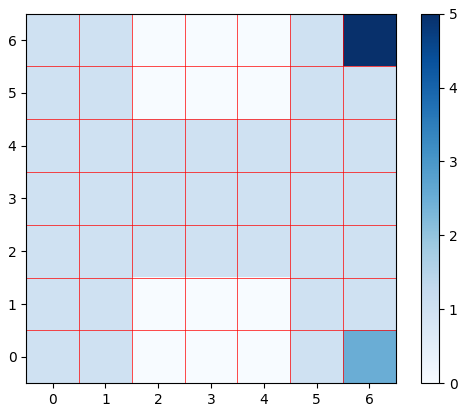}} \hfil
\subfloat[MCE-IRL Reward]{\includegraphics[scale = 0.25]{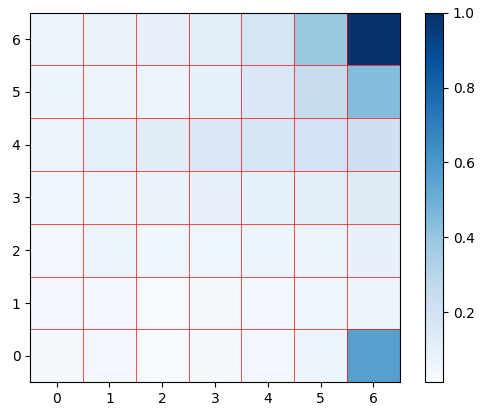}}
\caption{Results for $7 \times 7$ sequential goal grid-world.}
\label{fig:moirl_7x7}
\end{figure*}

\vspace*{-5pt}
The plots in \autoref{fig:moirl_7x7} show the demonstration and learned robot policy for the multi-goal $7 \times 7$ grid-world. Left figures in each sub-figure represent learned/inferred rewards. Right figures show the grid-world with start state (light blue), goal (dark blue), obstacles (red) and demonstration/policy (green). There are two goals and the rewards are inferred accordingly. At the next step, the algorithm enumerates all possible policies: (a) $start \rightarrow goal_1 \rightarrow goal_2$ and (b) $start \rightarrow goal_2 \rightarrow goal_1$. The final policy is a hybrid of the demonstrations while trying to minimize the time (soft requirement). In this case, it infers a policy $start \rightarrow goal_1$ (top-right) $\rightarrow goal_2$ (bottom-right). For sequential goals, MCE-IRL is unable to learn any reward even from 300 demonstrations. As we see in the figure, MCE-IRL has 2 problems: (1) it doesn't learn the reward for obstacles/avoid regions and (2) it learns only when there are 2 independent terminal states, i.e., it does not consider the sequential visitation of goals or that all goals must be covered by the policy. Hence a policy with the MCE-IRL reward and our multi-sequential goal algorithm is forced to visit $goal_2$ and then $goal_1$, thereby restricting the specification only to this order. MCE-IRL can also learn higher rewards for states other than the terminal states. Hence, for some cases in experiments, the policy was able to visit only one goal while ignoring the other.

\vspace*{-10pt}
\subsection{Frozenlake}
\begin{figure*}[htbp]
\centering
\subfloat[Demo 1]{\includegraphics[width= 1.7in]{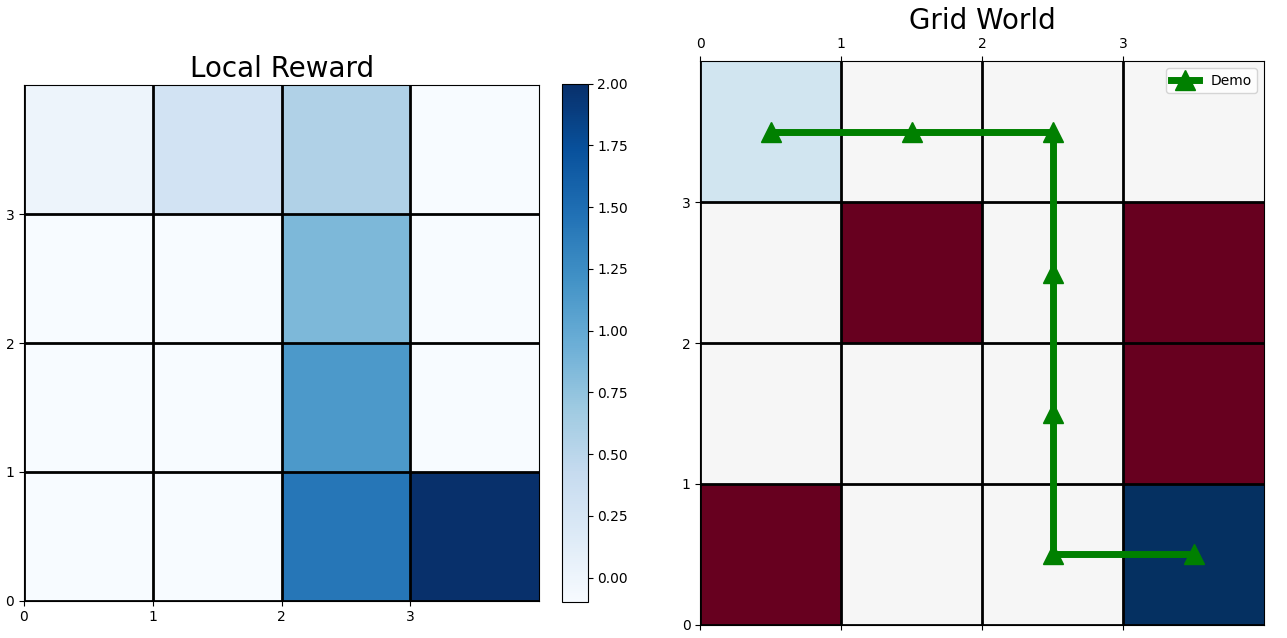}} \hfil
\subfloat[Demo 2]{\includegraphics[width= 1.7in]{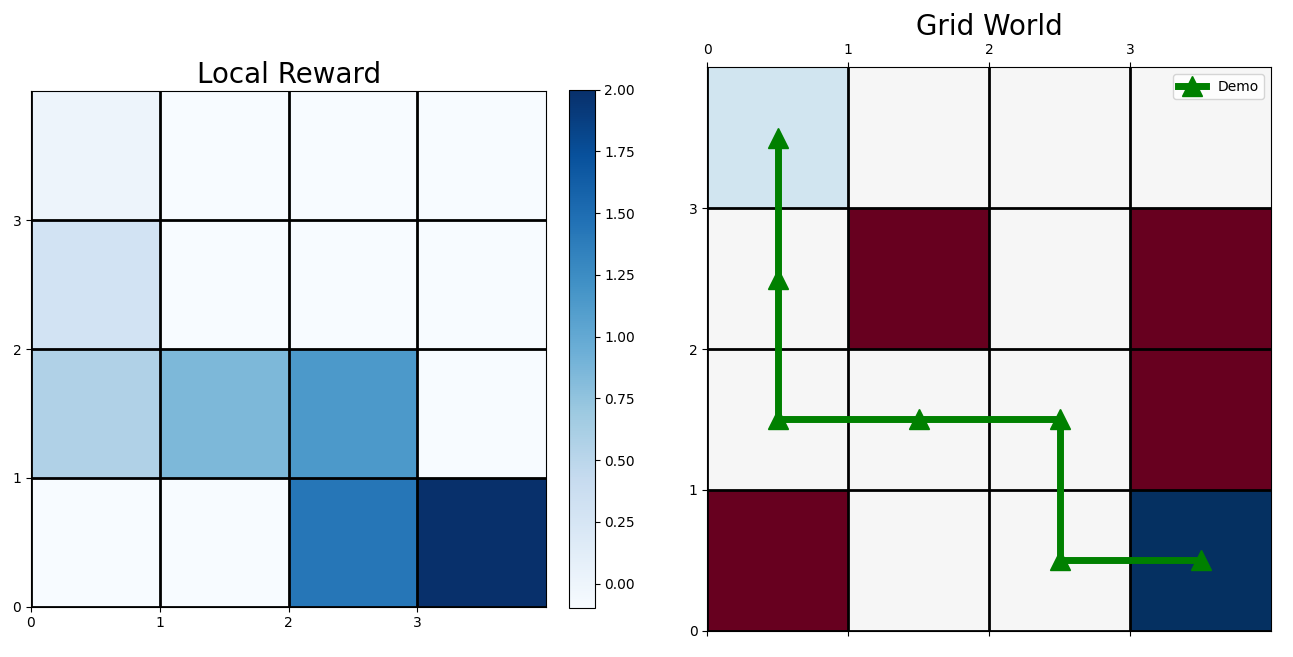}} \hfil
\subfloat[Robot Policy]{\includegraphics[width= 1.7in]{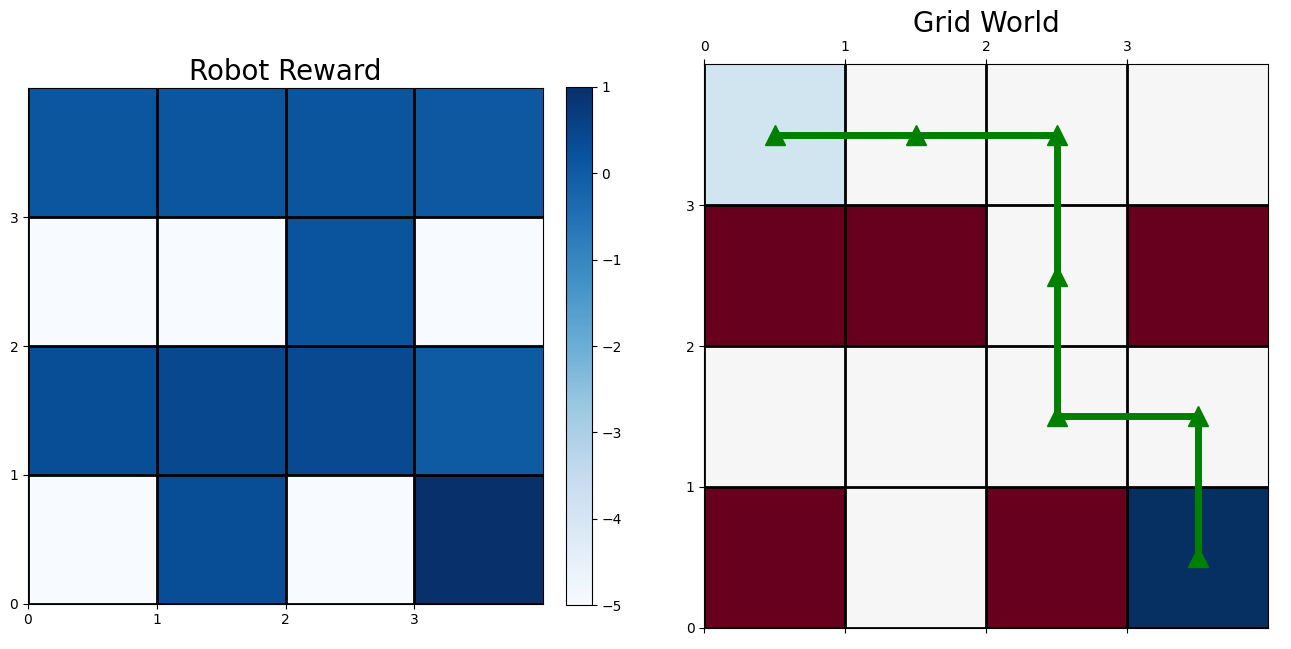}}

\caption{Results for $4 \times 4$ FrozenLake}
\label{fig:frozenlake}
\end{figure*}

\begin{figure}[htbp]
\centering
\subfloat[Standard Q-Learning $4 \times 4$ grid]{\includegraphics[width= 2.7in, height=3cm]{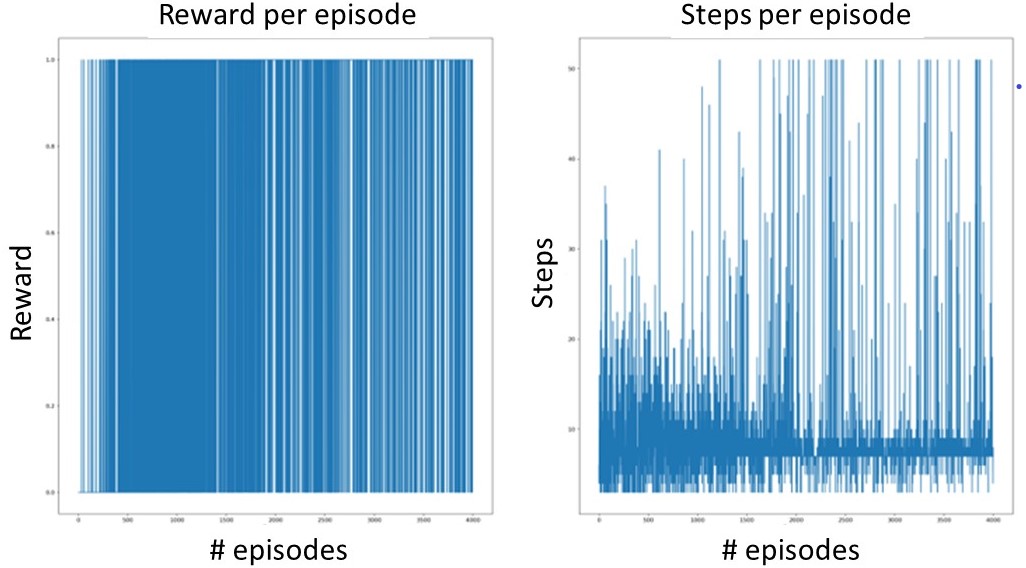}} \hfil
\subfloat[LfD+STL $4 \times 4$ grid]{\includegraphics[width= 2.7in, height=3cm]{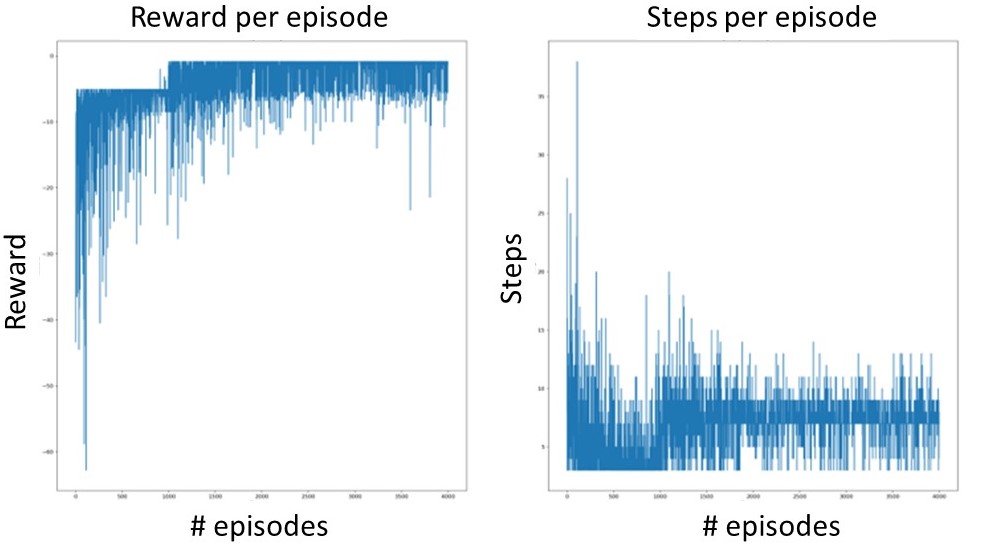}}
\caption{Statistics indicating the exploration rate of each algorithm as well as rewards accumulated in each training episode.}
\label{fig:frozenlake_stats}
\end{figure}

\vspace*{-10pt}
The results in \autoref{fig:frozenlake} show the robot policy in which demonstrations were provided on one map, but the agent had to use that information and explore on an unseen map. Left figures of each sub-figure represent learned rewards. Right figures show the grid-world with start state (light blue), goal (dark blue), obstacles or holes (red) and demonstration/policy (green). The robot is finally tested on a different map. The figures in \autoref{fig:frozenlake_stats} show comparisons in the exploration space between our method and standard Q-Learning with hand-crafted rewards. 
\vspace*{-10pt}
\vspace*{-10pt}

%
%

\begin{figure*}[htbp]
\centering
\subfloat[Demo 1]{\includegraphics[width= 2in]{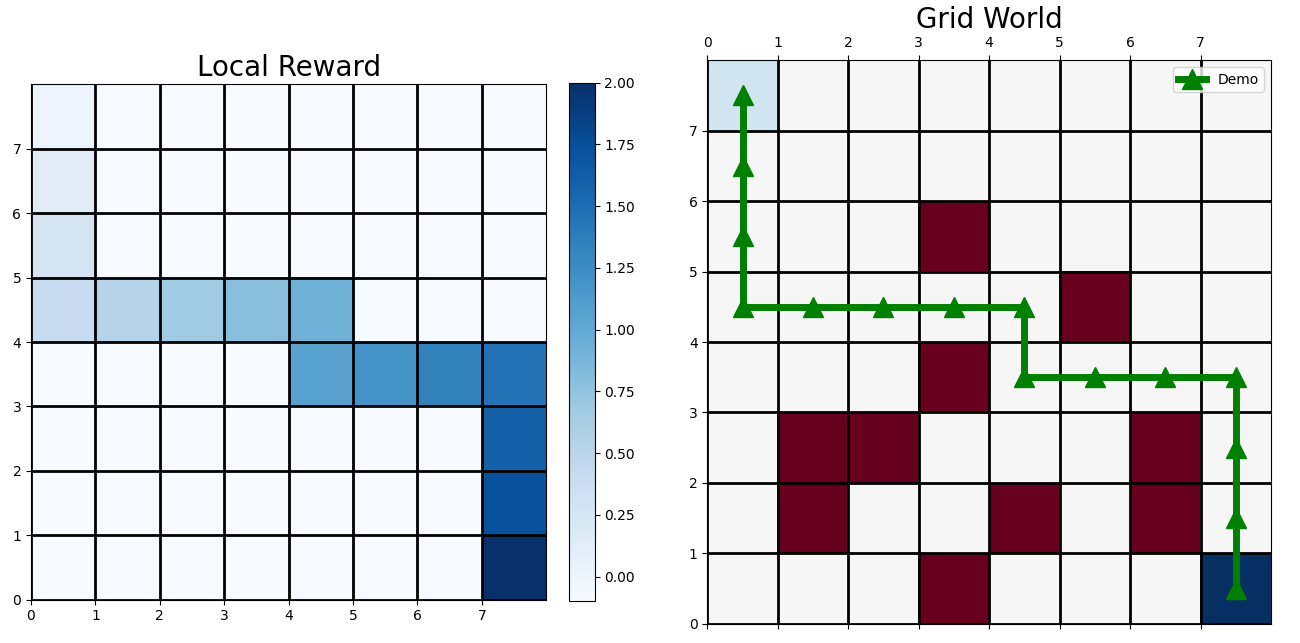}} \hfil
\subfloat[Demo 2]{\includegraphics[width= 2in]{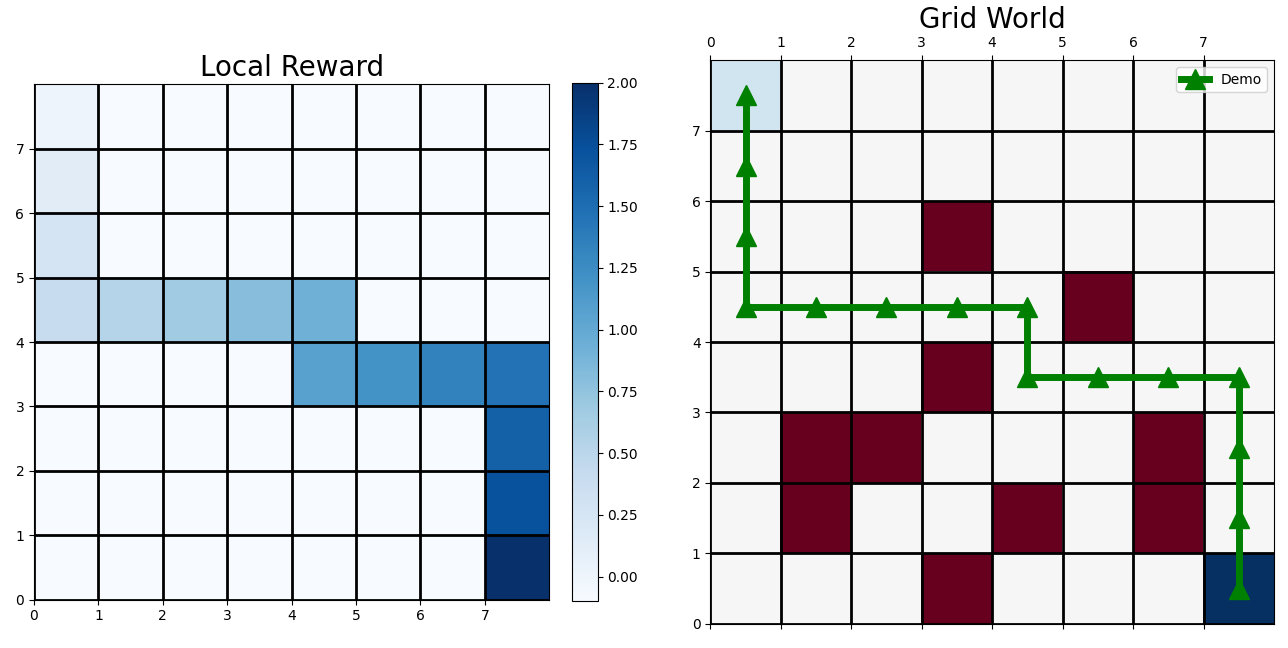}}

\subfloat[Robot Policy on test map 1]{\includegraphics[width= 1.8in]{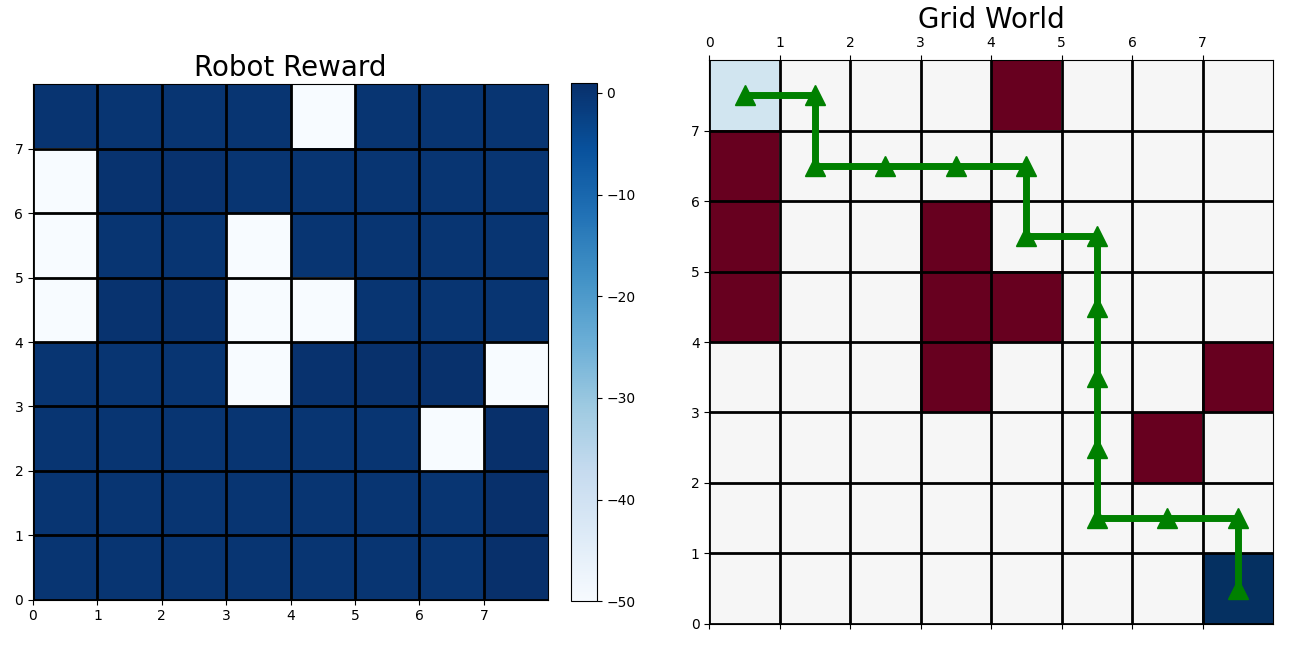}} \hfil
\subfloat[Robot Policy on test map 2]{\includegraphics[width= 1.8in]{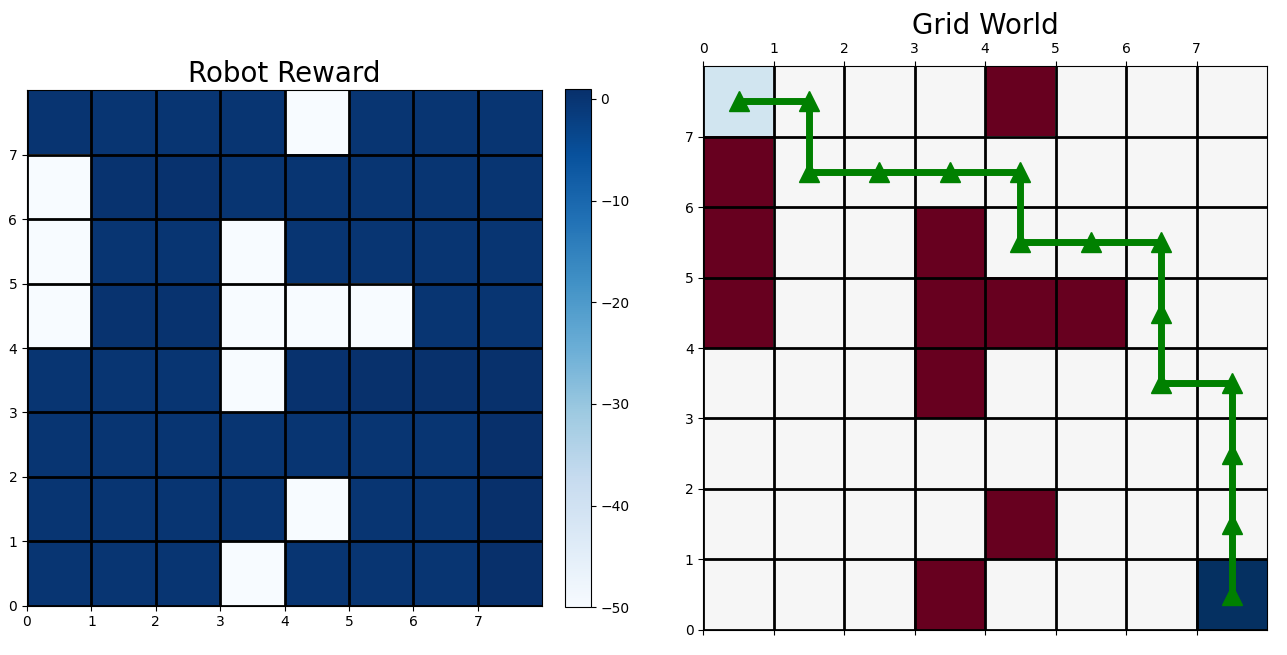}} \hfil
\subfloat[Robot Policy on test map 3]{\includegraphics[width= 1.8in]{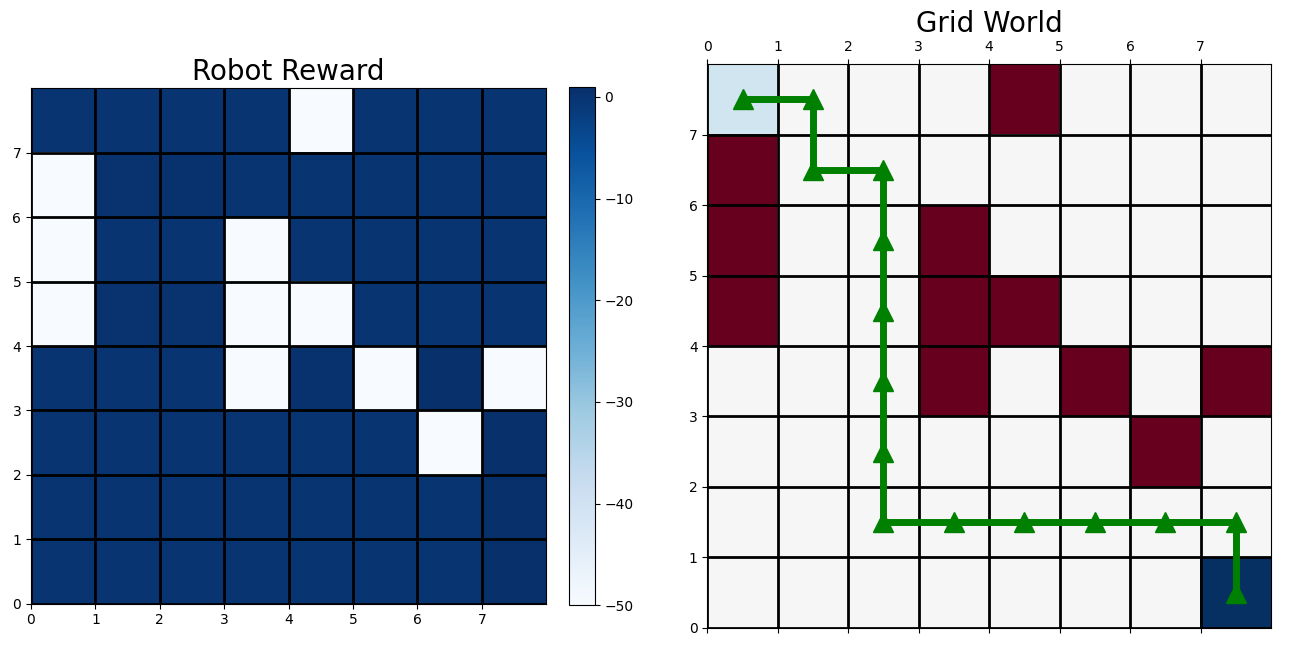}}
\caption{Results for $8 \times 8$ FrozenLake. Left subfigures represent the reward and the right subfigures show the environment and policy.}
\label{fig:frozenlake_grid8}
\end{figure*}

Similar results were obtained in the $8 \times 8$ grid size Frozenlake (see \autoref{fig:frozenlake_grid8}). A total of 5 demonstrations (4 good and 1 incomplete) were provided on a particular map. The agent then had to explore and learn a policy on 3 different maps using only the rewards from the map on which demonstrations were provided. The obstacles were moved about in each of the test/unseen maps and we see that the agent was able to successfully learn a policy to reach the goal.
\vspace*{-10pt}
\subsection{Mountain Car Results}
\begin{figure*}[htbp]
\centering
\subfloat[Demo 1]{\includegraphics[width= 2.8in]{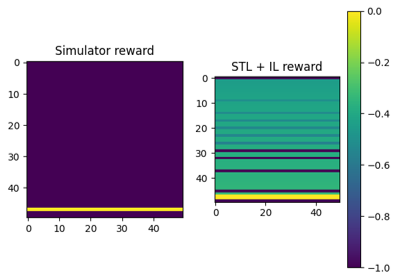}}
\subfloat[Demo 2]{\includegraphics[width= 2.8in]{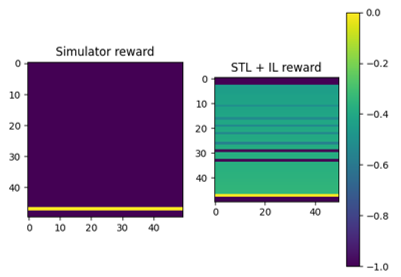}}

\subfloat[Final robot reward]{\includegraphics[width=1.5in, height=4cm]{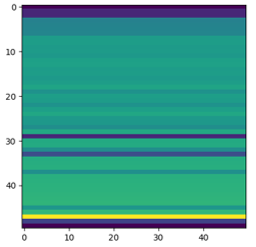}}
\caption{(a) and (b): The left figures represent the simulator reward (1 at goal and 0 elsewhere) while the right figures show the rewards based on STL specification. (c) Rewards inferred from demonstrations. \textit{Note: In all the figures, the axes represent the cell numbers corresponding to the grid size.}}
\label{fig:mountain_car_demo}
\end{figure*}

\vspace*{-10pt}
In a similar manner we show the demonstrations and rewards inferred in this environment. For the mountain car, we used a $50 \times 50$ grid size to show the scalability of our approach and its performance for sparse rewards (\autoref{fig:mountain_car_demo}). Other grid sizes used for the experiments were $75 \times 75$ and $100 \times 100$.
\vspace*{-15pt}

\end{document}